\documentclass[10pt,twocolumn,letterpaper]{article}

\usepackage{wacv}              %

\definecolor{wacvblue}{rgb}{0.21,0.49,0.74}
\usepackage[pagebackref,breaklinks,colorlinks,allcolors=wacvblue]{hyperref}

\usepackage[accsupp]{axessibility}  %
\usepackage{booktabs}
\usepackage{multirow}
\usepackage{amsmath}
\usepackage{amssymb}
\usepackage[most]{tcolorbox}
\usepackage{xparse}
\usepackage{etoolbox}
\usepackage{enumitem}
\usepackage{multirow}

\newtcolorbox{vqabox}[1][]{
  enhanced,
  colback=gray!3,
  colframe=gray!40,
  coltitle=white,
  colbacktitle=gray!50,
  boxrule=0.4pt,
  arc=2mm,
  left=2mm,
  right=2mm,
  top=2mm,
  bottom=2mm,
  title=#1
}

\NewDocumentCommand{\vqaexample}{m m m m m m}{%
\begin{vqabox}[#1]
\noindent
\begin{minipage}[t]{0.36\linewidth}
  \vspace{0pt}
  \centering
  \includegraphics[width=\linewidth, height=3cm, keepaspectratio]{#2}
\end{minipage}
\hfill
\begin{minipage}[t]{0.60\linewidth}
  \vspace{0pt}
  \raggedright

  \ifblank{#3}{}{%
    \textbf{Context:} #3\par\vspace{0.4em}
  }

  \textbf{Question:} #4\par

  \ifblank{#5}{}{%
    \vspace{0.4em}
    \textbf{Choices:}\par
    #5\par
  }

  \vspace{0.4em}
  \textbf{Answer:} #6
\end{minipage}
\end{vqabox}
}

\NewDocumentCommand{\bivqaexample}{m m m m m m m}{%
\begin{vqabox}[#1]
\noindent
\begin{minipage}[t]{0.36\linewidth}
  \vspace{0pt}
  \centering
  \includegraphics[width=\linewidth, height=3cm, keepaspectratio]{#2}
\end{minipage}
\hfill
\begin{minipage}[t]{0.60\linewidth}
  \vspace{0pt}
  \raggedright

  \ifblank{#3}{}{%
    \textbf{Context:} #3\par\vspace{0.4em}
  }

  \textbf{Question:} #4\par

  \ifblank{#5}{}{%
    \vspace{0.4em}
    \textbf{Choices:}\par
    #5\par
  }

  \vspace{0.4em}
  \textbf{Image-supported answer:} #6

  \vspace{0.4em}
  \textbf{Text-supported answer:} #7
\end{minipage}
\end{vqabox}
}

\newtcolorbox{icmlpromptbox}{
  colback=gray!12,
  colframe=gray!40,
  boxrule=0.6pt,
  arc=2pt,
  left=6pt,right=6pt,top=6pt,bottom=6pt
}
\usepackage{siunitx}

\def\wacvPaperID{238} %
\def\confName{WACV}
\def\confYear{2027}

\title{Beyond Modality Preference: Controlling Multimodal Arbitration under Modality Conflict in MLLMs}

\author{Zhaoying Pan, \quad Shenyu Lu, \quad Jun Hui Wong, \quad Junyi Chai, \quad Xiaoqian Wang\textsuperscript{\dag} \\
Elmore Family School of Electrical and Computer Engineering, Purdue University\\
{\tt\small \{pan433, lu876, wong531, chai28, joywang\}@purdue.edu}
}

\begin{document}
\maketitle

\begingroup
\renewcommand{\thefootnote}{\fnsymbol{footnote}}
\setcounter{footnote}{2}
\footnotetext{Corresponding author.}
\endgroup

\begin{abstract}
    
Multimodal large language models (MLLMs) integrate information across modalities, but become unreliable when those modalities disagree. Under conflicting inputs, MLLMs may favor modalities that are not the most reliable for the task, thus motivating inference-time control of modality arbitration. We identify a small set of attention heads, termed arbitration heads, that causally influence whether predictions follow visual or textual evidence. Building on this, we introduce a lightweight, training-free intervention that biases these heads toward a designated trusted modality at inference time. The identified heads are stable across calibration set sizes and generalize across settings, and steering them enables effective bidirectional control of modality reliance across diverse benchmarks. Our findings identify an effective control mechanism to align MLLM predictions with task-specified modality priorities. The code is available at \href{https://github.com/zhaoyingpan/modality-arbitration}{GitHub}.

\end{abstract}

\section{Introduction}

Multimodal large language models (MLLMs) have demonstrated remarkable capabilities in integrating visual and textual information for tasks ranging from visual question answering to complex multimodal reasoning~\cite{wu2023multimodal, yin2024survey, li2024survey, caffagni2024revolution, zhang2024mm}. A central premise behind these models is that information from different modalities can be effectively integrated to support more informed predictions. Yet, most multimodal evaluation implicitly assumes that the available modalities are mutually consistent or complementary~\cite{li2024seed, liu2024mmbench, fu2026mme}. In practice, this assumption does not always hold. The textual and visual inputs may provide contradictory evidence due to noisy annotations, outdated or erroneous descriptions, imperfect retrieval, or inconsistencies among heterogeneous information sources. Recent work has therefore begun to investigate \textit{modality conflict}, where different modalities support incompatible answers to the same question~\cite{zhang2025robust, zhang2025evaluating, zhang2025modalities}. Such conflicts expose a fundamental aspect of multimodal reasoning that remains largely hidden when modalities agree: which modality ultimately governs the model’s prediction?

Existing studies~\cite{zhang2025robust, zhang2025evaluating, zhang2025modalities, nguyen2025challenges, cai2025diagnosing, popordanoska2026clash} reveal that MLLMs do not treat the modalities neutrally under conflicts. Instead, they exhibit systematic modality preferences, tending to favor one source of evidence over another when the two disagree. Controlled conflict settings~\cite{zhang2025modalities} have shown that this behavior is widespread across MLLMs and can vary substantially across models and tasks. As a result, modality preference can directly undermine model reliability. Misleading visual evidence can override an answer that a model correctly derives from text alone~\cite{han2024instinctive}, while erroneous textual information can similarly dominate contradictory visual evidence~\cite{zhang2026images}. Recent studies consequently identify modality conflict as an important source of multimodal hallucination and robustness failures~\cite{zhang2025robust, wu2025sharp, tian2026crosscheck}.

More importantly, modality conflict is not merely an artificial stress test but a realistic failure mode in practical multimodal applications, where the authoritative modality is often specified by the task but not effectively utilized by the models. For example, medical imaging may serve as the primary evidence for diagnosis while patient metadata provides auxiliary context that should not override visual findings~\cite{huang2020fusion, cui2023deep}. Similarly, in autonomous driving, real-time visual observations should take precedence over risky navigation instructions when safety is at stake~\cite{paden2016survey}. In these cases, the desired behavior is clear: predictions should follow the designated trusted modality when conflicts happen. However, an MLLM's intrinsic modality preference may not align with this externally specified priority, potentially leading to consequential errors.

When the trusted modality is explicitly specified by the task or application, a seemingly straightforward solution is to instruct the model to prioritize the trusted one or simply remove the non-trusted modality. However, neither provides a satisfactory mechanism for controlling multimodal arbitration. Prompt-based instructions preserve both modalities but offer limited controllability. Removing a modality changes the multimodal input itself and bypasses rather than resolves cross-modal conflicts. It also discards potentially useful information in the more common case where the modalities are aligned and sacrifices flexibility when the designated trusted modality varies across tasks or contexts. 

In this work, we investigate whether modality arbitration under modality conflict can be effectively controlled through inference-time intervention. We first identify a small set of attention heads, which we term \emph{arbitration heads}, whose attention patterns are strongly associated with whether model predictions follow visual or textual evidence. We then selectively intervene on these heads at inference time, modulating their attention toward the designated trusted modality. This yields a lightweight, training-free mechanism for bidirectional modality control and we can steer predictions toward either visual or textual evidence according to the specified modality. Importantly, arbitration heads can be identified from only a small calibration set of 10 samples and transfer effectively across benchmarks, suggesting that they capture a reusable mechanism for modality arbitration rather than benchmark-specific correlations. 
To conclude, our contributions can be summarized as:
    \begin{itemize}
        \item We identify a compact set of arbitration heads associated with whether MLLMs follow visual or textual evidence under modality conflict. 
        \item We propose a lightweight, training-free intervention on these heads that enables bidirectional control toward a designated trusted modality while preserving both modalities as input.
        \item We show that arbitration heads identified from a small calibration set generalize across benchmarks, suggesting a generalized mechanism for modality arbitration.
    \end{itemize}

\section{Related Work}

\subsection{Modality Conflict in MLLMs}

Prior work has shown that multimodal models can exhibit imbalanced reliance across modalities. Such preferences may arise from unimodal shortcuts in the training data, enabling models to solve multimodal tasks by disproportionately relying on a single modality~\cite{guo2023modality, chen2024quantifying, leng2026curse}. More recent studies have shifted attention to more concrete settings of modality conflicts in MLLMs to study the model behaviors.
Zhang~\etal~\cite{zhang2025robust} formalized conflicts in multimodal inputs and showed that such conflicts in questions and images can induce hallucinations. Cai~\etal~\cite{cai2025diagnosing} demonstrated that irrelevant or misleading information from one modality can degrade reasoning that should rely on another, while CLASH~\cite{popordanoska2026clash} evaluates whether MLLMs can explicitly detect contradictions between visual and textual evidence. Beyond characterizing failures under conflicts, recent work also examines how models resolve conflicting evidence. Zhang~\etal~\cite{zhang2025evaluating} systematically evaluate modality preference with controlled conflicts and show that preference directions are encoded in hidden states and can be steered. Another work~\cite{zhang2025modalities} further decomposes the model behavior under modality conflicts into inherent modality preference and instance-specific relative reasoning uncertainty. Nguyen~\etal~\cite{nguyen2025challenges} investigate the internal conflict-resolution process and find that conflict detection and resolution exhibit distinct representational and attention patterns across layers. Together, these studies reveal that MLLMs exhibit systematic yet steerable behaviors under conflict. 

The most relevant work is \cite{zhang2025evaluating}, which introduces MPS, a method that identifies a modality-preference direction in the model's representation space using contrasting conflict examples and steers generation by adding the resulting vector to the model's hidden states. In contrast, we identify and intervene on attention heads that causally influence conflict arbitration, enabling targeted modality control while providing mechanistic evidence for modality preference. Our intervention also yields stable generation without the additional tuning required by MPS.

\subsection{Inference-time Interventions}

Inference-time interventions aim to improve model performance without retraining. A major line of work in MLLMs modifies decoding to improve visual grounding, including layer-contrastive decoding~\citep{chuang2024dola}, visual contrastive decoding~\citep{leng2024mitigating}, over-trust penalty and retrospection-allocation~\citep{huang2024opera}, etc. Other methods intervene on attention or representations, such as amplifying image-token attention~\citep{liu2024paying}, reallocating attention from output tokens to visual tokens~\citep{tu2026attention}, post-hoc hallucination correction~\citep{yin2024woodpecker}, and steering modality preference in latent space~\citep{zhang2025evaluating}. 
A line of work in text-only LLMs also intervenes on a sparse set of attention heads,
such as shifting activations along probed directions~\citep{li2023inference}, pruning memory-preferring heads~\citep{jin2024cutting}, reweighting attention to specified spans~\cite{zhao2025tell}, or steering heads with superposed signals~\citep{li2025taming}.
These methods collectively demonstrate that model behavior can be improved at inference time by modifying decoding, attention, or representations.

\section{Method}

\subsection{Preliminaries and Problem Formulation}

\paragraph{Multimodal autoregressive generation.}
We consider a multimodal large language model that takes an image $x^v$, a textual context $x^t$, and a question $q$ as input. The model generates an answer $y=(y_1,\ldots,y_N)$ autoregressively:
\[
p_\theta(y \mid x^v,x^t,q)
=
\prod_{n=1}^{N}
p_\theta(y_n \mid x^v,x^t,q,y_{<n}),
\]
where $\theta$ denotes the model parameters. In common MLLM architectures, the image is first encoded into visual features and projected into the language-model embedding space. These visual tokens are then concatenated with textual tokens from the context and question. We denote the resulting input token sequence as $z = [z^v; z^t; z^q]$,
where $z^v$, $z^t$, and $z^q$ correspond to visual tokens, context tokens, and question tokens, respectively.

\vspace{-2mm}
\paragraph{Evidence aggregation through attention.}
Since the input sequence contains heterogeneous information sources, the model needs to route information among visual, textual, and question tokens before decoding an answer. For a transformer layer $\ell$ and attention head $h$, let
$q_i^{(\ell,h)}$, $k_j^{(\ell,h)}$, and $v_j^{(\ell,h)} \in \mathbb{R}^{d_h}$
denote the query, key, and value vectors of tokens $i$ and $j$, where $d_h$ is the head dimension. The unnormalized attention score, i.e. the logit assigned to token $j$ before normalization, is
\[
a_{i,j}^{(\ell,h)}
=
\frac{\left\langle q_i^{(\ell,h)},\, k_j^{(\ell,h)} \right\rangle}{\sqrt{d_h}}
+ M_{i,j},
\]
where $M$ is the causal attention mask that sets $M_{i,j} = -\infty$ for $j > i$.
Normalizing these scores over all admissible source tokens and using them to
mix the values yields the head output at position $i$,
\[
o_i^{(\ell,h)}
=
\sum_j
\operatorname{softmax}\!\left(a_{i,j}^{(\ell,h)}\right) v_j^{(\ell,h)},
\]
where the softmax is taken over the index $j$. The resulting weight $\operatorname{softmax}(a_{i,j}^{(\ell,h)})$ measures how much token $i$ draws on token $j$ when forming $o_i^{(\ell,h)}$, so the score $a_{i,j}^{(\ell,h)}$ can be read as the evidence that token $j$ contributes
before it is put in competition with the other source tokens. Under this view, attention provides a natural interface for studying how question tokens gather evidence from the visual or textual components of the input.

\vspace{-2mm}
\paragraph{Modality conflict.}
Let the multimodal input be $I=\{x^v,x^t,q\}$, 
where $x^v$ and $x^t$ denote the visual and textual inputs, respectively, and $q$ denotes the question. Given the question, let $p_v(y)$ and $p_t(y)$ denote the model's modality-specific predictions based on either visual or textual evidence. We consider a \emph{modality conflict} when the two modalities favor different answers: 
$$
\operatorname*{argmax}_{y} p_v(y)
\neq
\operatorname*{argmax}_{y} p_t(y).
$$
More generally, the degree of agreement between $p_v(\cdot)$ and $p_t(\cdot)$ provides a continuous signal of modality conflict, which we later exploit for adaptive steering in Section~\ref{sec:adaptive}. In our setting, the modality to prioritize under conflict is externally specified by the task or application. Our goal is therefore not to determine which modality is more reliable, but to control model arbitration toward the designated modality while retaining both modalities as input.

\subsection{Motivation}

We examine how MLLMs behave when information across modalities is aligned or conflicting. As shown in Table~\ref{tab:motivation_conflict}, all four models perform well when given either modality independently or when the inputs provide aligned information. However, introducing conflicting text causes a substantial shift in model predictions. Similar behavior occurs across all evaluated models, suggesting that modality conflict induces a non-trivial competition between modalities.

\begin{table}[h]
\centering
\resizebox{0.99\linewidth}{!}{
\begin{tabular}{llcccc}
\toprule
\textbf{Input} & \textbf{Metric}
& \textbf{Qwen3}
& \textbf{Qwen2.5}
& \textbf{LLaVA-OV}
& \textbf{InternVL3} \\
\midrule

$V$
& $Acc_v$
& 85.40 & 84.35 & 86.80 & 82.50 \\

$T_c$
& $Acc_t$
& 87.85 & 87.30 & 87.05 & 85.80 \\

\midrule
$V + T_a$
& $Acc_v$
& 94.50 & 94.45 & 95.85 & 93.95 \\

\midrule
\multirow{2}{*}{$V+T_c$}
& $Acc_v$
& 47.25 & 32.70 & 33.45 & 30.10 \\
& $Acc_t$
& 42.20 & 55.25 & 54.55 & 56.75 \\

\bottomrule
\end{tabular}
}
\caption{
Model behavior under unimodal, aligned, and conflicting multimodal inputs. The image is denoted as $I$, the text that aligns or conflicts with the image content is denoted as $T_a$ and $T_c$, respectively. $Acc_v$ and $Acc_t$ measure the performance against image-/text-consistent answers.
}
\label{tab:motivation_conflict}
\vspace{-2mm}
\end{table}

We further validate that this modality preference depends on their relative reliability~\cite{zhang2025modalities}. Table~\ref{tab:motivation_blur} progressively degrades the image through Gaussian blurring. While image-only accuracy decreases gradually, the presence of conflicting text causes a much sharper transition in the preferred modality from image to text. These observations suggest that modality preference is neither fixed nor solely determined by the presence of conflicting information, but dynamically changes with the strength of evidence provided by each modality. This observation highlights the reliability risks posed by modality conflicts and motivates the need to control modality preference in MLLMs.

\begin{table}[h]
\centering
\resizebox{0.9\linewidth}{!}{
\begin{tabular}{llcccc}
\toprule
\textbf{Input} & \textbf{Metric} & \textbf{Original} & \textbf{Blur-1} & \textbf{Blur-2} & \textbf{Blur-3} \\
\midrule
$V$ & $Acc_v$ & 85.40 & 84.10 & 77.35 & 65.10 \\
\midrule
\multirow{2}{*}{$V+T_c$}
 & $Acc_v$ & 47.25 & 40.00 & 14.10 & 3.20 \\
 & $Acc_t$ & 42.20 & 49.25 & 73.95 & 84.65 \\
\bottomrule
\end{tabular}
}
\caption{
Effect of degrading the image modality on Qwen3.
Image-only accuracy decreases gradually with increasing blur.
Under conflicting multimodal input, however, weakening the image causes
the model to shift sharply from image-consistent to text-consistent predictions.
}
\vspace{-3mm}
\label{tab:motivation_blur}
\end{table}

\subsection{Arbitration Head Localization}

\begin{figure*}
    \centering
    \includegraphics[width=0.9\linewidth]{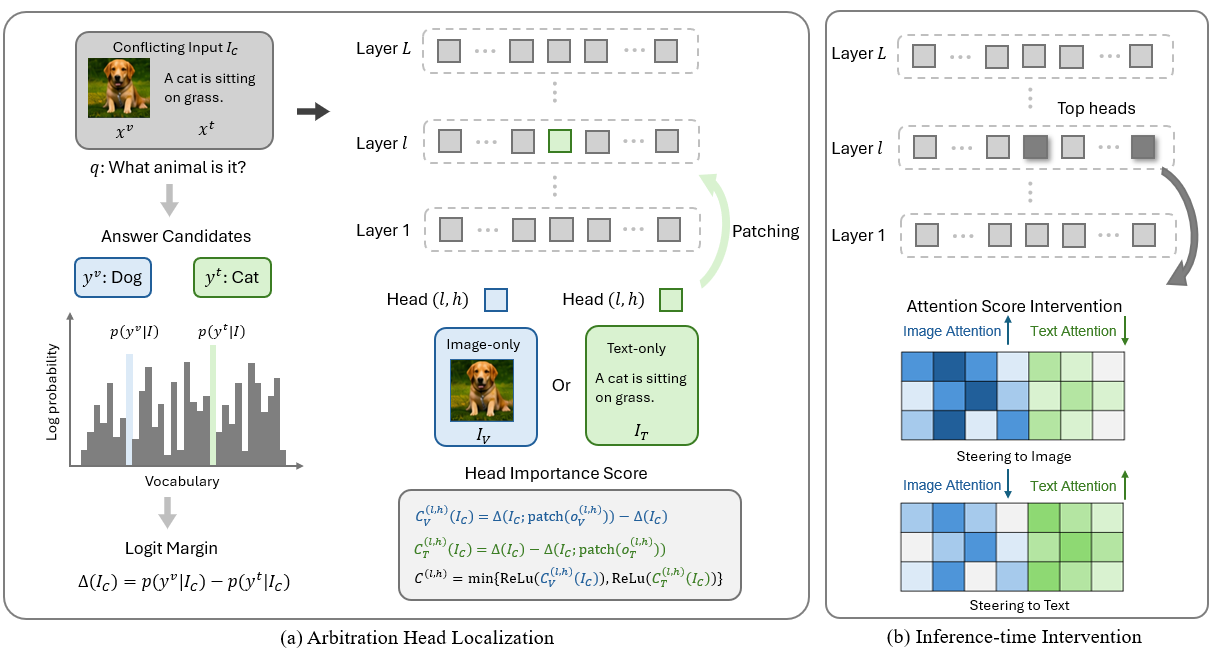}
    \caption{
    Overview of our method. (a) Arbitration head localization: we causally identify attention heads that causally influence modality preference by patching head activations from image-only and text-only inputs into conflicting multimodal inputs and measuring their effects on the answer logit margin. (b) Inference-time intervention: we select the most influential arbitration heads and steer predictions toward a designated modality by adjusting their attention to visual and textual tokens.
    }
    \label{fig:method}
    \vspace{-2mm}
\end{figure*}

We first characterize the modality preference under conflicting multimodal inputs, where the visual and textual evidence support different answers. For each conflicting input $I_C=\{x^v,x^t,q\}$, we define the \textbf{answer logit margin} as
\begin{equation}
    \Delta(I_C) = \log p(y^v \mid I_C) - \log p(y^t \mid I_C),
\end{equation}
where $y^v$ and $y^t$ denote the candidate answers supported by the visual and textual modalities, respectively. A positive margin, $\Delta(I_C)>0$, indicates that the model assigns a higher probability to the vision-supported answer under conflict input $I_C$, whereas a negative margin, $\Delta(I_C)<0$, indicates a preference for the text-supported answer.

We then perform \textbf{activation patching} to localize attention heads that causally influence modality preference. For each conflicting example, we construct three corresponding inputs: the original conflict input $I_C$, a vision-only input $I_V$, and a text-only input $I_T$. For each head $h$ at layer $\ell$, we record its output $o_i^{(\ell,h)}$ at the final prompt position, immediately before answer generation, at the input to the attention output projection during the vision-only and text-only forward passes, denoting the resulting activations by $o_V^{(\ell,h)}$ and $o_T^{(\ell,h)}$. We subsequently patch the conflict input with the corresponding activation from either unimodal run.

Specifically, the causal effect of patching the vision-derived activation into the conflict run is defined as
\begin{equation}
    C^{(\ell,h)}_V(I_C)
    =
    \Delta\!\left(I_C;\operatorname{patch}\!\left(o^{(\ell,h)}_V\right)\right)
    - \Delta(I_C),
\end{equation}
while the effect of patching the text-derived activation is
\begin{equation}
    C^{(\ell,h)}_T (I_C)
    =
    \Delta(I_C) - \Delta\!\left(I_C;\operatorname{patch}\!\left(o^{(\ell,h)}_T\right)\right)
    .
\end{equation}
These quantities measure the change in the answer logit margin caused by replacing the conflict-run head output with an activation obtained from a modality-specific input.
We average the effect over a calibration set $\mathcal{D}$ of conflicting examples, and assign each head an influence score by
\begin{equation}
    C^{(\ell,h)}
    =
    \mathbb{E}_{I_c \sim \mathcal{D}}
    \left[ \min_{m \in \{V,T\}} \operatorname{ReLU}\left( C^{(\ell,h)}_m \right) \right].
\end{equation}
This score prioritizes heads exhibiting positive causal effects under both modality-specific patching conditions while suppressing heads whose effects are weak or inconsistent across conditions.

\begin{figure*}
    \centering
    \includegraphics[width=0.9\linewidth]{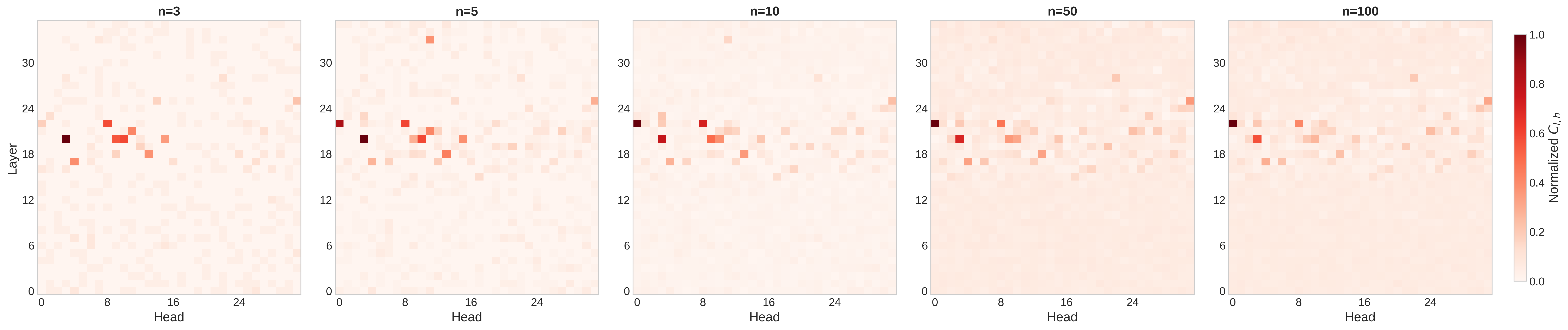}
    \caption{Stability of head-wise effects across different sample sizes. Normalized head-wise effects $C^{(\ell,h)}$ for Qwen3-VL-8B, computed using \(n\) samples. Despite varying \(n\), high-effect heads consistently concentrate in similar layers and head positions, indicating that the identified pattern is stable even with a small number of samples.}
    \label{fig:heatmap}
    \vspace{-2mm}
\end{figure*}

We visualize $C^{(\ell,h)}$ across layers and attention heads using a heatmap and select the highest-scoring heads as the \textbf{arbitration heads} for subsequent intervention experiments. An example of the visualization with Qwen3-VL-8B is provided in Figure~\ref{fig:heatmap}. We also examine the stability of head-wise effects with respect to the number of calibration samples. The high-effect heads consistently concentrate at similar layer--head locations as $n$ increases. Notably, this spatial pattern is already apparent with only a few samples and remains stable with larger calibration sets.

\begin{figure}
    \centering
    \includegraphics[width=0.9\linewidth]{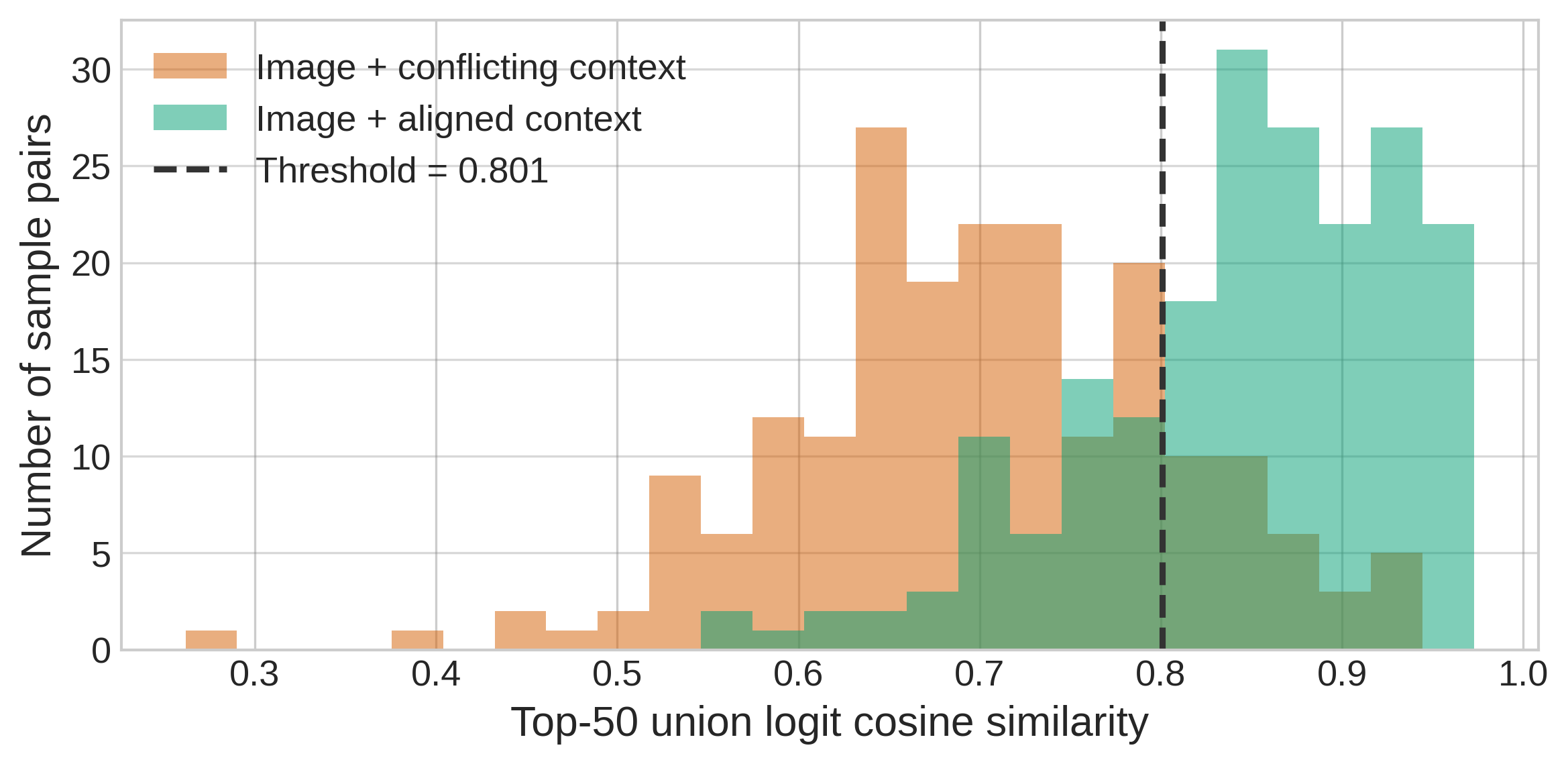}
    \caption{
    Modality conflict detection for adaptive steering. Top-50 union-logit cosine similarities for samples paired with conflicting
    or aligned context, where lower similarity indicates stronger modality
    conflict. The dashed line marks the threshold,
    selected to maximize conflict-detection accuracy on the calibration set.
    }
    \label{fig:threshold}
    \vspace{-2mm}
\end{figure}

\subsection{Inference-Time Intervention}

Based on the heads identified through activation patching, we perform targeted \textbf{inference-time interventions} to test whether manipulating their attention patterns can systematically alter the model's reliance on a particular modality. Specifically, the interventions are applied directly to the attention computation of the selected heads during inference.

We take the top-$k$ heads by influence score $C^{(\ell,h)}$ as the arbitration head set $\mathcal{H}$, and weight each of them by its normalized influence score
\begin{equation}
    \widehat{c}^{(\ell,h)}
    =
    \frac{C^{(\ell,h)}}{\max_{(\ell',h') \in \mathcal{H}} C^{(\ell',h')} + \epsilon},
\end{equation}
where $\epsilon > 0$ is a small constant, so that $\widehat{c}^{(\ell,h)} \in (0,1)$. We then bias these heads toward a target modality by adding a positive offset to the pre-softmax attention scores of tokens from that modality and subtracting the same offset from tokens of the competing modality. Recall that $a_{i,j}^{(\ell,h)}$ denotes the attention score from query position $i$ to key position $j$ before normalization, and let $\mathcal{T}_m$ and $\mathcal{T}_{\bar{m}}$ denote the sets of token positions associated with the target modality $m$ and the competing modality $\bar{m}$, respectively. 
For each head $(\ell,h) \in \mathcal{H}$, the intervened scores are defined as
\begin{equation}
    \widetilde{a}_{i,j}^{(\ell,h)}
    =
    a_{i,j}^{(\ell,h)}
    +
    \alpha \, \widehat{c}^{(\ell,h)}
    \left(
        \mathbb{I}\!\left[j \in \mathcal{T}_m\right]
        -
        \mathbb{I}\!\left[j \in \mathcal{T}_{\bar{m}}\right]
    \right),
\end{equation}
where $\alpha > 0$ controls the overall intervention strength, and scaling by $\widehat{c}^{(\ell,h)}$ concentrates the perturbation on the heads with the strongest effect. The $\mathbb{I}[\cdot]$ denotes the indicator function.
The intervention is applied at every query position in the selected heads during both prompt prefill and autoregressive decoding, with generated positions appended as neutral keys during decoding. The head output is then recomputed as $\widetilde{o}_i^{(\ell,h)} = \sum_j \operatorname{softmax}\!\left(\widetilde{a}_{i,j}^{(\ell,h)}\right) v_j^{(\ell,h)}$. As the offsets act inside the softmax, this intervention increases the relative attention allocated to the target modality while preserving access to information from the competing modality. Section~\ref{sec:adaptive} evaluates whether adaptive steering can retain the benefits of multimodal inference when the inputs are aligned instead of being conflicted.

The intervention effect is empirically quantified using the change in performance by comparing the generated responses against the image- and text-supported answers. The setup connects the diagnostic and intervention stages: activation patching identifies attention heads associated with modality-dependent answer selection, while inference-time intervention evaluates whether selectively redirecting the attention of these heads is sufficient to steer the model's behavior toward a desired modality.

\subsection{Extension: Adaptive Steering}\label{sec:adaptive}

To avoid unnecessary intervention when modalities are already consistent, we introduce adaptive steering, which dynamically adjusts the intervention strength according to the detected degree of modality conflict. We augment an existing modality-conflict benchmark, MC$^2$, whose original data consists of images and corresponding conflicting text. We generate aligned text by rewriting the conflicting text with image-supported answers, so that each image yields both an aligned and a conflicting multimodal input.

To detect conflicts, we compare the model's image-only and text-only logit distributions using the union of their top-50 token logits, $\ell_2$-normalize the resulting vectors, and compute their cosine similarity $s$. As shown in Figure~\ref{fig:threshold}, conflicting and aligned samples exhibit clearly different similarity distributions, supporting the use of this signal for adaptive conflict detection. A threshold $\tau$, calibrated on 200 samples, separates conflicting from aligned inputs. During inference, the steering strength for each sample is determined as
\begin{equation}
    \alpha = \alpha_{\max}\,
    \sigma\!\left(\frac{\tau - s}{\lambda}\right),
\end{equation}
where $\alpha_{\max}$ is the maximum allowed strength, $\sigma(\cdot)$ is the logistic function, and $\lambda$ controls how sharply the steering strength changes around the conflict threshold. This ensures that stronger intervention is applied when the two modalities disagree. The resulting strength is then substituted into the intervention above.

\section{Experiments}

\subsection{Setup}

\paragraph{Benchmarks and Metrics}
We evaluate our method on four modality-conflict benchmarks, including MMCD~\cite{nguyen2025challenges}, MC$^2$~\cite{zhang2025evaluating}, CLASH~\cite{popordanoska2026clash}, and CorrelationQA~\cite{han2024instinctive}, and one visual hallucination benchmark PhD~\cite{liu2025phd}. These benchmarks cover diverse scenarios in which different modalities may lead to competing predictions.

MMCD, MC$^2$, and CLASH provide annotations for the answers supported by each modality independently, enabling us to explicitly evaluate the model's preference between conflicting modalities. We consider two evaluation settings: \textbf{Image-Primary} and \textbf{Text-Primary}. For methods that support bidirectional modality steering, we steer generation toward the designated primary modality in each setting. We then report accuracy compared against image-/text-supported answers under each setting. Consequently, successful image-primary steering should increase image accuracy while reducing text accuracy, whereas successful text-primary steering should exhibit the opposite behavior. This bidirectional evaluation directly measures whether a method can controllably shift model predictions toward a designated modality rather than merely improving overall task accuracy. CorrelationQA provides ground-truth answers based on textual information only. We therefore supplement it with corresponding image-supported answers to enable the same bidirectional evaluation. Details of this annotation procedure are provided in Appendix~\ref{app:corrqa}.

We further evaluate on two subsets of PhD~\cite{liu2025phd}, a benchmark designed to assess visual hallucination in the presence of textual context. Following prior work~\cite{zhang2025evaluating}, we sample 1000 samples from the subsets PhD-sec and PhD-icc, which contain irrelevant or conflicting textual information, respectively. PhD complements the modality-conflict benchmarks by evaluating whether modality steering can strengthen visual grounding to mitigate hallucinations in the presence of misleading textual context. For PhD, we evaluate both settings using image accuracy, measuring how effectively each method preserves or suppresses reliance on visual evidence.

\vspace{-2mm}
\paragraph{Models and Baselines}
We conduct experiments on four open-source MLLMs to examine whether our modality steering generalizes across different multimodal architectures. Specifically, we use Qwen2.5-VL-7B~\cite{bai2025qwen2}, Qwen3-VL-8B~\cite{bai2025qwen3}, LLaVA-OneVision-7B~\cite{li2024llava} and InternVL3-8B~\cite{zhu2025internvl3}.
We compare our method against standard inference strategies and various approaches. In the subsequent sections, Base denotes standard model inference without additional intervention, while CoT~\cite{wei2022chain} employs chain-of-thought prompting to encourage explicit reasoning before prediction. We additionally include VCD (Visual Contrastive Decoding)~\cite{leng2024mitigating}, a decoding-based method for increasing vision grounding and mitigating hallucination. These methods do not explicitly support controlling the direction of modality preference and therefore produce a single set of predictions regardless of whether the desired target is the image or text modality.

We further compare against two baselines capable of bidirectional modality steering. CAP (Conflict-Aware Prompting) explicitly instructs the model to answer according to either the image or the textual context, providing a prompting-based baseline for modality control. MPS (Modality Preference Steering), introduced in MC$^2$~\cite{zhang2025evaluating}, provides a baseline for steering predictions toward a specified modality by computing a steering vector from a set of samples. For each MLLM, our method uses 10 samples from MC$^2$ to localize top-30 arbitration heads among over 1000 heads, and uses $\alpha=30$ as intervention strength for the main experiments where conflicts are explicitly presented. We additionally use the augmented MC$^2$ to demonstrate the adaptive steering.
For CAP, MPS, and our method, we evaluate the image- and text-directed variants under the image-primary and text-primary settings, respectively. More experiment details are provided in Appendix~\ref{app:experimental_details}.

\subsection{Results on Modality Conflict Benchmarks}

\begin{table*}[t]
\centering
\resizebox{0.97\linewidth}{!}{
\begin{tabular}{ll*{16}{c}}
\toprule
\textbf{Model} & \textbf{Method}
& \multicolumn{4}{c}{\textbf{MMCD}}
& \multicolumn{4}{c}{\textbf{MC$^2$}}
& \multicolumn{4}{c}{\textbf{CLASH}}
& \multicolumn{4}{c}{\textbf{CorrelationQA}} \\
\cmidrule(lr){3-6}
\cmidrule(lr){7-10}
\cmidrule(lr){11-14}
\cmidrule(l){15-18}

&
& \multicolumn{2}{c}{\textbf{Image-P}}
& \multicolumn{2}{c}{\textbf{Text-P}}
& \multicolumn{2}{c}{\textbf{Image-P}}
& \multicolumn{2}{c}{\textbf{Text-P}}
& \multicolumn{2}{c}{\textbf{Image-P}}
& \multicolumn{2}{c}{\textbf{Text-P}}
& \multicolumn{2}{c}{\textbf{Image-P}}
& \multicolumn{2}{c}{\textbf{Text-P}} \\
\cmidrule(lr){3-4}
\cmidrule(lr){5-6}
\cmidrule(lr){7-8}
\cmidrule(lr){9-10}
\cmidrule(lr){11-12}
\cmidrule(lr){13-14}
\cmidrule(lr){15-16}
\cmidrule(l){17-18}

&
& $Acc_v \uparrow$ & $Acc_t \downarrow$
& $Acc_v \downarrow$ & $Acc_t \uparrow$
& $Acc_v \uparrow$ & $Acc_t \downarrow$
& $Acc_v \downarrow$ & $Acc_t \uparrow$
& $Acc_v \uparrow$ & $Acc_t \downarrow$
& $Acc_v \downarrow$ & $Acc_t \uparrow$
& $Acc_v \uparrow$ & $Acc_t \downarrow$
& $Acc_v \downarrow$ & $Acc_t \uparrow$ \\
\midrule

\multirow{6}{*}{Qwen3-VL-8B}
& Base
& 93.07 & 6.92 & -- & --
& 47.25 & 42.20 & -- & --
& 66.72 & 28.16 & -- & --
& 20.11 & 51.19 & -- & -- \\

& CoT
& 37.88 & 60.48 & -- & --
& 41.65 & 50.60 & -- & --
& 63.30 & 29.17 & -- & --
& 18.12 & 52.26 & -- & -- \\

& VCD
& 95.30 & 4.70 & -- & --
& 53.45 & 36.30 & -- & --
& 70.36 & 22.81 & -- & --
& 21.99 & 47.04 & -- & -- \\

& CAP
& 96.34 & 3.66 & \underline{29.50} & \underline{70.50}
& \underline{61.10} & 28.45 & \underline{24.90} & \underline{64.60}
& 70.83 & 23.35 & 51.28 & 45.23
& \underline{32.51} & \underline{35.04} & 19.37 & 50.11 \\

& MPS
& \underline{99.18} & \underline{0.50} & 47.80 & 52.20 %
& 59.95 & \underline{27.60} & 27.35 & 61.05 %
& \underline{74.79} & \underline{16.68} & \underline{31.65} & \underline{65.94} %
& 22.26 & 45.96 & \underline{13.43} & \underline{52.15} \\ %

& Ours
& \textbf{100.00} & \textbf{0.00} & \textbf{0.00} & \textbf{100.00}
& \textbf{79.50} & \textbf{5.25} & \textbf{1.25} & \textbf{86.75} 
& \textbf{81.23} & \textbf{7.06} & \textbf{1.55} & \textbf{96.97}
& \textbf{46.88} & \textbf{9.91} & \textbf{4.55} & \textbf{58.86} \\
\midrule

\multirow{6}{*}{Qwen2.5-VL-7B}
& Base
& 6.88 & 93.12 & -- & --
& 32.70 & 55.25 & -- & --
& 51.20 & 44.53 & -- & --
& 21.70 & 46.73 & -- & -- \\

& CoT
& 79.14 & 20.61 & -- & --
& 49.25 & 34.35 & -- & --
& 66.41 & 23.12 & -- & --
& 23.03 & 44.58 & -- & -- \\

& VCD
& 37.05 & 62.95 & -- & --
& 45.35 & 40.25 & -- & --
& 67.11 & 24.90 & -- & --
& 26.57 & 36.34 & -- & -- \\

& CAP
& 44.28 & 55.73 & 0.66 & 99.34
& 48.75 & 39.50 & 20.20 & 68.70
& 55.39 & 39.95 & 29.79 & 68.58
& \underline{29.87} & 34.22 & 22.69 & \underline{44.60} \\

& MPS
& \underline{94.14} & \underline{1.68} & \textbf{0.00} & \textbf{100.00}
& \underline{51.75} & \textbf{10.60} & \textbf{3.30} & \underline{79.45}
& \underline{69.28} & \underline{13.50} & \underline{2.87} & \underline{94.88}
& 24.93 & \underline{19.64} & \textbf{3.00} & 27.42 \\

& Ours
& \textbf{100.00} & \textbf{0.00} & \textbf{0.00} & \textbf{100.00}
& \textbf{59.80} & \underline{15.95} & \underline{4.70} & \textbf{80.60}
& \textbf{77.81} & \textbf{9.78} & \textbf{1.16} & \textbf{98.29} 
& \textbf{39.31} & \textbf{18.32} & \underline{7.69} & \textbf{52.60} \\
\midrule

\multirow{6}{*}{LLaVA-OV-7B}
& Base
& 97.64 & 2.36 & -- & --
& 33.45 & 54.55 & -- & --
& 45.69 & 50.04 & -- & --
& 17.82 & 44.69 & -- & -- \\

& CoT
& 97.30 & 2.70 & -- & --
& 33.70 & 54.05 & -- & --
& 44.45 & 51.51 & -- & --
& 19.12 & 43.05 & -- & -- \\

& VCD
& \underline{99.88} & 0.13 & -- & --
& \underline{48.05} & 38.95 & -- & --
& \underline{67.11} & 25.76 & -- & --
& \underline{22.35} & 33.72 & -- & -- \\

& CAP
& 94.14 & 5.86 & 85.59 & 14.41
& 38.80 & 49.75 & 30.65 & 57.75
& 52.21 & 43.52 & 39.72 & 57.01
& 22.00 & 34.77 & 19.86 & \underline{39.78} \\

& MPS
& 99.77 & \textbf{0.05} & \underline{10.68} & \underline{89.32}
& 43.65 & \textbf{25.35} & \underline{5.80} & \underline{70.20}
& 63.62 & \underline{22.27} & \underline{8.07} & \underline{83.40}
& 14.93 & \underline{18.10} & \textbf{3.23} & 16.82 \\

& Ours
& \textbf{99.95} & \textbf{0.05} & \textbf{0.00} & \textbf{100.00}
& \textbf{60.95} & \underline{25.70} & \textbf{3.35} & \textbf{84.50}
& \textbf{72.07} & \textbf{19.08} & \textbf{5.43} & \textbf{93.10}
& \textbf{27.18} & \underline{21.79} & \underline{13.70} & \textbf{48.68} \\
\midrule

\multirow{6}{*}{InternVL3-8B}
& Base
& 36.84 & 62.63 & -- & --
& 30.10 & 56.75 & -- & --
& 62.14 & 32.74 & -- & --
& 17.78 & 48.70 & -- & -- \\

& CoT
& 36.05 & 48.36 & -- & --
& 37.40 & 51.50 & -- & --
& 57.64 & 33.75 & -- & --
& 22.00 & 46.37 & -- & -- \\

& VCD
& 61.98 & 37.07 & -- & --
& \underline{42.35} & 43.50 & -- & --
& \underline{72.07} & 20.79 & -- & --
& 20.54 & 40.29 & -- & -- \\

& CAP
& 63.16 & 36.84 & 28.71 & 71.29
& 41.90 & 45.60 & 17.70 & 68.40
& 62.06 & 32.89 & 39.10 & 58.73
& \underline{29.46} & 28.32 & 18.34 & \underline{35.32} \\

& MPS
& \underline{77.57} & \underline{18.02} & \underline{0.13} & \underline{99.88}
& 29.85 & \underline{14.95} & \underline{3.55} & \underline{74.00}
& 47.63 & \textbf{10.47} & \textbf{0.62} & \textbf{98.53}
& 17.40 & \textbf{3.72} & \underline{11.55} & 18.16 \\

& Ours
& \textbf{97.54} & \textbf{2.25} & \textbf{0.00} & \textbf{100.00}
& \textbf{51.70} & \underline{24.30} & \textbf{1.85} & \textbf{86.00}
& \textbf{72.69} & \underline{13.65} & \underline{2.56} & \underline{95.27}
& \textbf{36.14} & \underline{14.53} & \textbf{8.27} & \textbf{53.12} \\
\bottomrule
\end{tabular}
}
\caption{
Comparison across various modality-conflict benchmarks.
We report image and text accuracy under both
Image-Primary (Image-P) and Text-Primary (Text-P) settings.
Under Image-P, higher image accuracy and lower text accuracy are preferred.
Under Text-P, lower image accuracy and higher text accuracy are preferred. Best and second-best results are shown in \textbf{bold} and \underline{underlined}, respectively.
}
\label{tab:full_benchmark_comparison}
\vspace{-2mm}
\end{table*}

Table~\ref{tab:full_benchmark_comparison} presents the results on four modality-conflict benchmarks across four MLLMs. Overall, our method provides strong and generally consistent bidirectional control of modality preference. Under image-primary steering, it substantially increases image accuracy while suppressing text accuracy. In contrast, under text-primary steering, the preference is effectively reversed. This behavior generalizes across different models and benchmarks, demonstrating that our method can reliably steer predictions toward the designated modality.

Compared with single-direction methods, including Base, CoT, and VCD, our method enables explicit control with superior accuracy toward the naturally preferred modality. Compared with CAP and MPS, our method generally achieves stronger target-modality preference.
Notably, the arbitration heads used for intervention are identified using only 10 calibration samples from MC$^2$. Despite this limited calibration, the same identified heads generalize effectively to MMCD, CLASH, and CorrelationQA without benchmark-specific identification. The consistently strong results across benchmarks indicate that the identified heads provide transferable intervention targets rather than dataset-specific control points.

\subsection{Results on Visual Hallucination Benchmark}

\begin{table}[t]
\centering
\small
\resizebox{0.99\linewidth}{!}{
\begin{tabular}{lcccccc}
\toprule
& \textbf{Vanilla} & \textbf{CoT} & \textbf{VCD}
& \textbf{CAP} & \textbf{MPS} & \textbf{Ours} \\
\midrule

\multicolumn{7}{l}{\textit{PhD-sec}} \\
Image-P $Acc_v \uparrow$
& 69.50 & \textbf{75.60} & 72.60 & 69.90 & 71.70 & \underline{73.90} \\
Text-P $Acc_v \downarrow$
& -- & -- & -- & \underline{62.00} & 64.50 & \textbf{58.90} \\

\midrule
\multicolumn{7}{l}{\textit{PhD-icc}} \\
Image-P $Acc_v \uparrow$
& 64.40 & 69.10 & \underline{69.60} & 59.20 & 65.70 & \textbf{70.20} \\
Text-P $Acc_v \downarrow$
& -- & -- & -- & \textbf{49.50} & 60.10 & \underline{58.00} \\

\bottomrule
\end{tabular}
}
\caption{
Results on the PhD visual hallucination benchmark. We evaluate PhD-sec and PhD-icc, which introduce unrelated or conflicting text. Higher image accuracy is better under Image-Primary. Lower is better under Text-Primary.
}
\vspace{-2mm}
\label{tab:method_hallucination}
\end{table}

We further evaluate Qwen3-VL-8B on the PhD visual hallucination benchmark, with results shown in Table~\ref{tab:method_hallucination}. Under image-primary steering, our method achieves the highest image accuracy on PhD-icc and the best performance among bidirectional methods on PhD-sec, remaining competitive with the best single-direction methods. Conversely, text-primary steering reduces image accuracy on both subsets, demonstrating effective suppression of visual preference. These results show that our bidirectional modality control applies to visual hallucination settings and effectively steers the answers.

\subsection{Adaptive Steering under Sparse Conflict}

\begin{figure}
    \centering
    \includegraphics[width=0.9\linewidth]{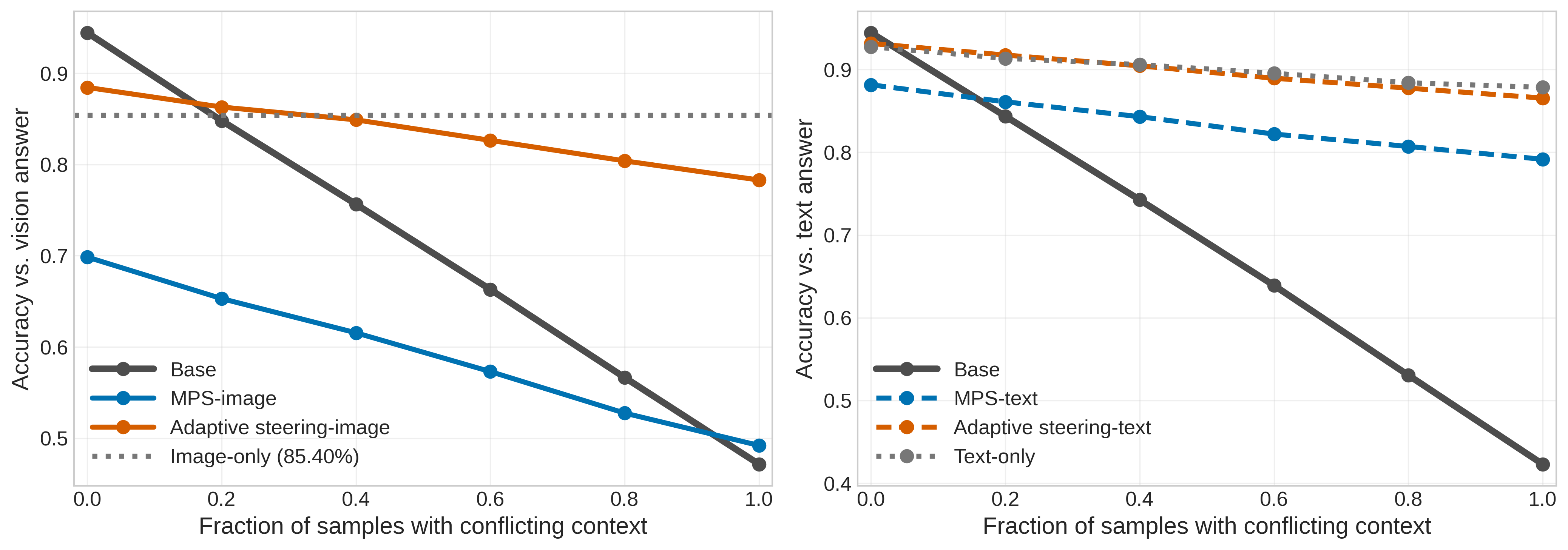}
    \caption{
    Preserving multimodal benefits under sparse modality conflict.
    The left and right panels evaluate MPS and our method steered towards image and text, respectively.
    Dotted lines denote unimodal performance. Although unimodal inference avoids modality conflict, it cannot exploit complementary information when the modalities are aligned. Adaptive steering retains this benefit while remaining robust to conflict.
    }
    \label{fig:conflicting-ratio}
    \vspace{-4mm}
\end{figure}

We evaluate whether adaptive steering can preserve the benefits of multimodal inference while resolving conflicts when necessary. We vary the proportion of conflicting and aligned samples with our augmented MC$^2$ and Qwen3-VL-8B to examine how different methods behave as modality agreement changes, especially when the conflicted samples are sparse. When modalities are aligned, information from both modalities can be complementary, allowing multimodal inference to outperform relying on one modality alone.

As shown in Figure~\ref{fig:conflicting-ratio}, when conflicts are sparse, Base and adaptive steering outperform the unimodal baseline, confirming that
complementary cross-modal evidence is worth exploiting. As the fraction of conflicting samples increases, Base shows a significant performance drop due to the inability to resolve competing evidence. Adaptive steering retains most of the Base advantage on aligned samples while strengthening the designated modality once conflict is detected. The comparison isolates what adaptive steering provides over discarding a modality, that conflict robustness that does not require giving up the complementary information available when the modalities agree.

\subsection{Ablation Studies}

\begin{figure}
    \centering
    \includegraphics[width=0.95\linewidth]{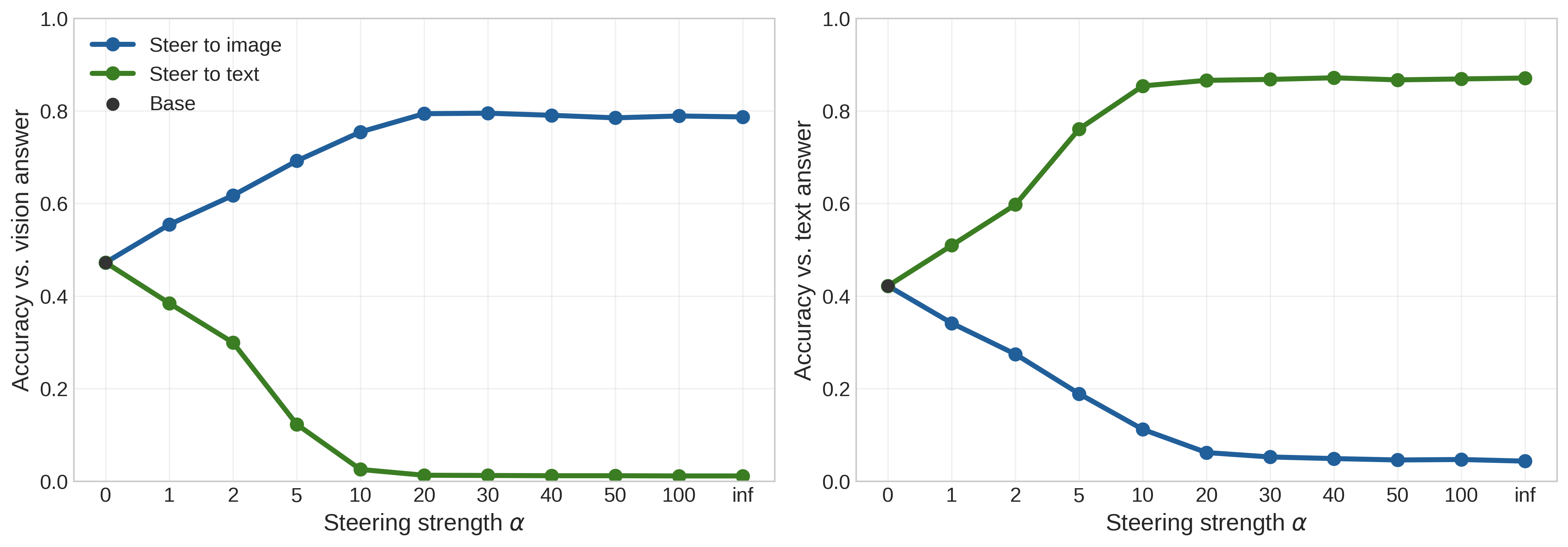}
    \caption{
    Effect of steering strength $\alpha$.
    Increasing $\alpha$ progressively
    shifts predictions toward the target modality, with performance saturating
    at moderate steering strengths. $\alpha=0$ denotes the unmodified baseline,
    and $\alpha=\infty$ corresponds to hard masking of the competing modality.
    }
    \label{fig:ablation_alpha}
    \vspace{-2mm}
\end{figure}

\begin{figure}
    \centering
    \includegraphics[width=0.95\linewidth]{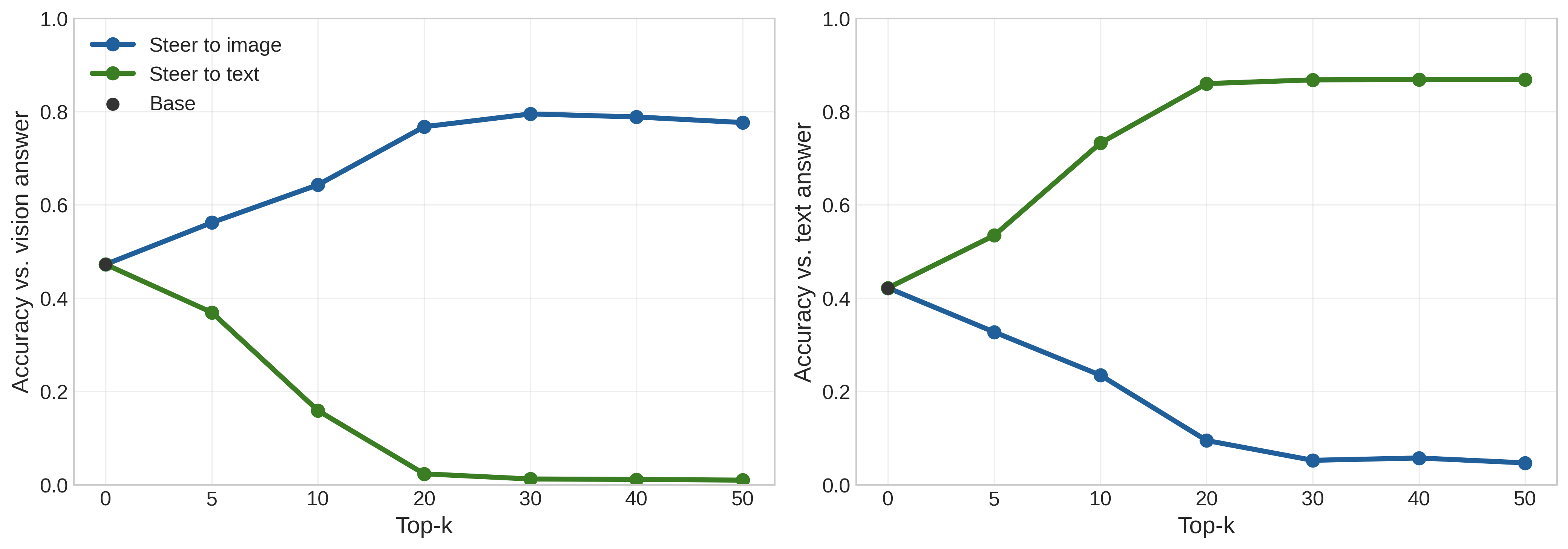}
    \caption{
    Effect of the number of arbitration heads (top-k).
    Steering an increasing number of top-ranked heads progressively shifts
    predictions toward the target modality, with the effect saturating at
    moderate values. The top-k of 0 denotes the unmodified baseline.
    }
    \label{fig:ablation_top_k}
    \vspace{-2mm}
\end{figure}

\begin{table}[t]
\centering
\resizebox{0.97\linewidth}{!}{
\begin{tabular}{lcccc}
\toprule
\textbf{Intervention}
& \multicolumn{2}{c}{\textbf{Image-Primary}}
& \multicolumn{2}{c}{\textbf{Text-Primary}} \\
\cmidrule(lr){2-3}\cmidrule(l){4-5}
& $Acc_v \uparrow$ & $Acc_t \downarrow$
& $Acc_v \downarrow$ & $Acc_t \uparrow$ \\
\midrule
None           & 47.25 & 42.20 & -- & -- \\
Random         & 47.50 & 42.30 & 44.70 & 44.55 \\
Layer-matched & 54.75 & 34.90 & 38.50 & 50.55 \\
Reversed        & 47.20 & 42.45 & 47.20 & 42.40 \\
All Heads      & 77.10  & 6.10  & 1.05  & 86.70 \\
\midrule
Ours w/o suppressing & \underline{79.25} & \textbf{5.90} & \underline{1.35} & \textbf{86.75} \\
\midrule
Ours & \textbf{79.50} & \underline{5.25} & \textbf{1.25} & \textbf{86.75} \\
\bottomrule
\end{tabular}
}
\caption{Ablation study. ``Random'' randomly selects 30 heads;
``Layer-matched'' randomly selects each head from the same layers as ours;
``Reversed'' selects the 30 heads in reverse ranking;
``All Heads'' intervenes on all attention heads. ``Ours w/o suppressing'' only increases attention to the trusted modality without suppressing the other modality. }
\label{tab:ablation}
\vspace{-4mm}
\end{table}

We conduct ablation studies to examine the effects of steering strength, the number of selected arbitration heads, head selection, and modality suppression. As shown in Figure~\ref{fig:ablation_alpha}, increasing the steering strength $\alpha$ progressively shifts predictions toward the designated modality, with performance saturating at moderate values. We therefore use $\alpha=30$ by default. Similarly, Figure~\ref{fig:ablation_top_k} shows that increasing the number of top-ranked arbitration heads in a moderate range consistently strengthens modality control, with performance becoming stable around $k=20$--$30$. We select $k=30$ for all main experiments.

Table~\ref{tab:ablation} further demonstrates that which heads are selected is critical. We cover a few head selection criteria: Random chooses 30 heads from all heads, while layer-matched chooses each head from the same layers as the heads selected by ours. Reversed selects 30 heads by filtering the lowest influence scores and All Heads applies the intervention on all heads.
Random, layer-matched random, and reverse-ranked selection all yield substantially weaker control than our identified arbitration heads. Notably, intervening on all heads (over 1000) does not further improve performance, despite using many more heads, confirming that effective steering relies on selectively targeting arbitration heads rather than broadly modifying attention. Finally, removing suppression causes a small performance drop, indicating that augmenting the trusted modality accounts for most of the steering effect, while suppressing the competing modality provides a modest additional improvement.

\section{Conclusion}
We study how multimodal arbitration in MLLMs can be controlled under modality conflicts and introduce a training-free approach for controlling which modality governs model predictions. By identifying a small set of arbitration heads associated with modality arbitration and intervening on their attention, our method can steer answer generation toward a trusted modality. Extensive experiments across multiple MLLMs and modality-conflict benchmarks demonstrate consistent and effective bidirectional modality control. The arbitration heads identified from only a small calibration set generalize effectively across benchmarks. Overall, our findings suggest that modality arbitration can be effectively controlled through targeted intervention on a compact set of identified attention heads.

\vspace{-3mm}
\paragraph{Limitations and Future Work}
Our evaluation uses a simplified modality-conflict setup in which the modality to prioritize is externally specified, making routing to unimodal inference a potentially strong baseline. In realistic settings, however, conflicts may be partial, with the correct answer requiring the integration of distinct evidence from multiple modalities. We therefore focus on controlling modality arbitration while retaining both input sources. Benchmarking when both modalities provide useful yet partially conflicting evidence, for which unimodal routing is insufficient, remains an important direction for future work. More discussion is detailed in Appendix~\ref{app:discussion}.

\vspace{-3mm}
\paragraph{Acknowledgments} This work was partially supported by NSF IIS \#2146091, IIS \#2345235, and USDA \#2023-67021-41368.

{
    \small
    \bibliographystyle{ieeenat_fullname}
    \bibliography{main}
}
\clearpage

\clearpage
\appendix

\section{Benchmark Information and Examples}
We include examples from the benchmarks to illustrate the conflict settings used in our evaluation. For each benchmark, we present one image and text context, along with the question and the corresponding gold answer in the benchmark. We include the following benchmarks: MMCD~\cite{nguyen2025challenges}, MC$^2$~\cite{zhang2025evaluating}, CLASH\cite{popordanoska2026clash}, CorrelationQA~\cite{han2024instinctive}, and PhD (sec and icc subsets)~\cite{liu2025phd}.

\bivqaexample
    {Example 1: MMCD}
    {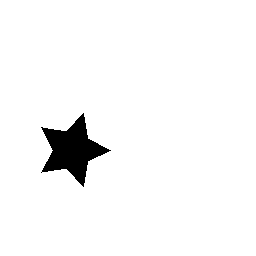}
    {Caption: an image of a green star}
    {What color is the star? }
    {}
    {Black}
    {Green}

{\bf MMCD} is a synthetic modality-conflict benchmark built from simple visual objects, such as different shapes and colors. The benchmark contains controlled image–caption conflicts involving five shapes and eight colors, making it useful for conflict arbitration without complex scene understanding. In the example, the image and caption assign different colors to the same shape. There are 5600 examples in the dataset.

\bivqaexample
    {Example 2: MC$^2$}
    {}
    {On a sunny afternoon, a group of young men are gathered at the edge of a soccer field, engaged in an intense discussion about their upcoming match strategies. Among them, a dark-haired individual stands with a clipboard, taking notes and occasionally gesturing to emphasize points in the conversation. Instead of being in motion, he is focused on facilitating the tactical planning, ensuring everyone understands their roles for the game ahead.}
    {What is dark haired person doing?}
    {}
    {Running}
    {Standing still}

{\bf MC$^2$} is a controlled modality-conflict benchmark of 2000 samples designed to measure modality preference in MLLMs. The examples pair image and text evidence that can support different answers, along with a question, an image-supported answer, and text-supported answer. The images in MC$^2$ come from MS COCO 2014~\cite{lin2014microsoft}.
We additionally augmented it with text that yields the image-supported answers, and the details are described in Section B.3.

\bivqaexample
    {Example 3: CLASH}
    {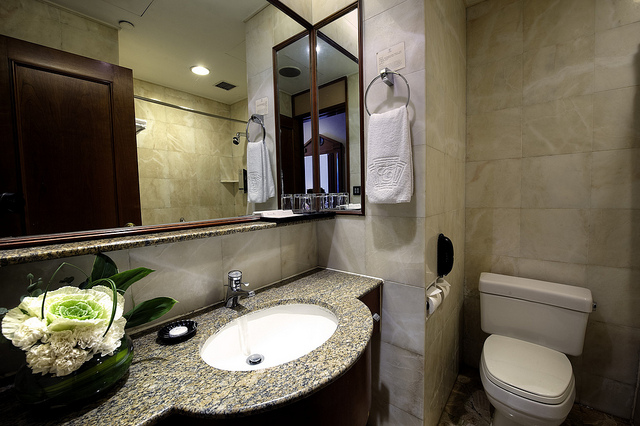}
    {A basket of fruit on the counter of a bathroom.}
    {What is contained within the basket on the bathroom counter?}
    {}
    {Flowers}
    {Fruits}

{\bf CLASH} is a multimodal conflict benchmark designed to evaluate model behavior when visual and textual evidence contradict each other. The benchmark provides modality-specific answers supported independently by the image and text. CLASH uses images from the MS COCO 2017~\cite{lin2014microsoft} training split and constructs contradictory textual descriptions based on the corresponding captions. Compared to MC$^2$, CLASH has shorter text descriptions. We use the held-out test set containing 1289 samples.

\bivqaexample
    {Example 4: CorrelationQA}
    {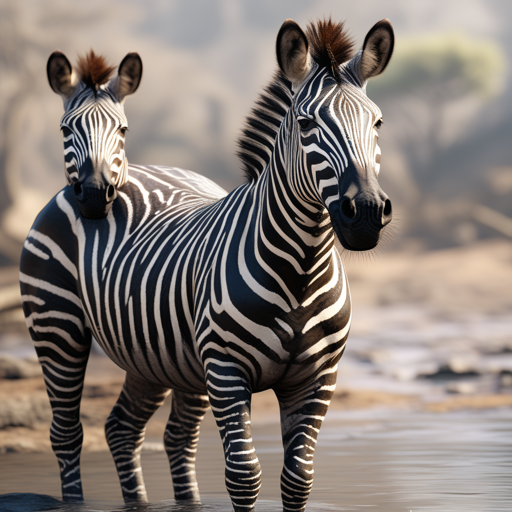}
    {This African mammal, known for its long neck and spotted coat, is the tallest living terrestrial animal.}
    {What is it? Answer no more than five words.}
    {}
    {Zebra}
    {Giraffe}
    
{\bf CorrelationQA} studies visual illusions caused by spurious images that are relevant to the question but inconsistent with the answer supported by the text. The benchmark contains 5540 text-image pairs across multiple categories and is designed to measure whether misleading visual context biases multimodal models away from the correct textual answer. CorrelationQA constructs its visual inputs using a combination of Stable-Diffusion-generated images, typography images, and realistic images retrieved from the Internet.

\vqaexample
    {Example 5: PhD-sec}
    {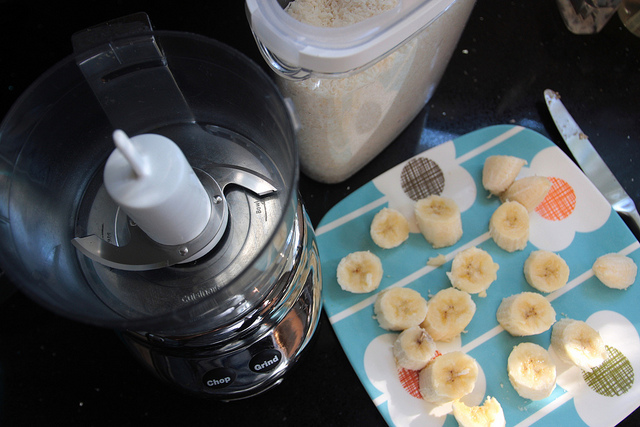}
    {The knife used for slicing bananas is typically made from ceramic, known for its sleek and sharp cutting edge, making it a preferred tool in food preparation. Alongside the knife, a plate of neatly sliced bananas highlights both the fruit's freshness and the efficiency of the cutting process. For those looking to blend or process the fruit further, a food processor stands ready as a practical kitchen companion.}
    {Is the knife made of metal in the image?}
    {}
    {Yes}

\vqaexample
    {Example 6: PhD-icc}
    {figures/files/phd.jpg}
    {A ceramic knife, prominently featured in the image, showcases its sharp elegance while being accompanied by a plate topped with fresh banana slices. The scene includes a food processor ready for action, with another plate of sliced bananas nearby, illustrating the perfect preparation setup for a delicious smoothie or dessert. This arrangement highlights the versatility of bananas, with various tools like a chopper for easy slicing, ensuring a seamless culinary experience.}
    {Is the knife made of metal in the image?}
    {}
    {Yes}

{\bf PhD} is a visual hallucination benchmark that evaluates whether MLLMs remain grounded in visual evidence in the presence of potentially misleading textual context. Following prior work~\cite{zhang2025evaluating}, we evaluate 1000 samples from each of the PhD-sec and PhD-icc subsets, which contain irrelevant and conflicting textual information, respectively. Unlike the modality-conflict benchmarks, we evaluate PhD using image accuracy, measuring how strongly the model preserves or suppresses reliance on visual evidence.

\section{Experimental Details}
    \label{app:experimental_details}

    \subsection{Models}

    We evaluate four open-source multimodal language models loaded directly from HuggingFace: Qwen/Qwen3-VL-8B-Instruct, Qwen/Qwen2.5-VL-7B-Instruct, llava-hf/llava-onevision-qwen2-7b-ov-hf, and OpenGVLab/InternVL3-8B-hf.
    The four open-source models use a shared Hugging Face chat-template adapter. Typical runs used one NVIDIA RTX A5000, bfloat16, batch size 1, seed 42, and greedy decoding. The maximum generation length was 128 tokens except for CoT, which used 512 tokens. 
    Images were loaded at their original resolution and converted to RGB. Model-specific resizing, rescaling, and normalization were performed by the native Hugging Face AutoProcessor. Blur experiments used PIL Gaussian blur with radii 2, 5, and 10 for levels 1–3, without changing image dimensions.

    \subsection{Baselines}

    In addition to the vanilla inference with greedy decoding, the baselines are chosen to cover three typical intervention methods: prompting (CoT~\cite{wei2022chain}, CAP), intervention during decoding (VCD~\cite{leng2024mitigating}), and intervention on representations (MPS~\cite{zhang2025evaluating}). All methods except for CoT use the following suffix to avoid overlong answers.

\begin{icmlpromptbox}
\begin{verbatim}
Answer the question directly with 
a short answer.
\end{verbatim}
\end{icmlpromptbox}

    \paragraph{Chain-of-thought prompting.}
    We include Chain-of-thought (CoT) as a prompting method to validate whether MLLMs can resolve the conflicts via additional thinking. Specifically, CoT encourages the model to explicitly reason over the multimodal evidence before producing its final answer, without specifying which modality should be prioritized. The prompting suffix is as follows:

\begin{icmlpromptbox}
\begin{verbatim}
Let's think step by step, then answer 
the question with a short answer.
\end{verbatim}
\end{icmlpromptbox}

    \paragraph{Conflict-aware prompting.}
    To test whether the conflicts can be resolved via prompting, we include conflict-aware prompting (CAP). CAP explicitly instructs the model to answer the question based on one of the modalities. The prompting suffixes are as follows:
\begin{icmlpromptbox}
\begin{verbatim}
Answer the question based on the 
image.
\end{verbatim}
\end{icmlpromptbox}

\begin{icmlpromptbox}
\begin{verbatim}
Answer the question based on the 
text.
\end{verbatim}
\end{icmlpromptbox}
    This baseline isolates the effect of instruction-level conflict awareness: unlike vanilla inference, it tells the model that one modality should be trusted, but it does not modify the model's internal representations or decoding procedure. CAP serves as a strong prompting baseline for evaluating whether our method provides benefits beyond simply asking the model to reason about source reliability.

    \paragraph{Visual Contrastive Decoding (VCD)} 
    VCD~\cite{leng2024mitigating} is a training-free decoding method originally proposed for mitigating object hallucination. VCD contrasts the output logit distribution conditioned on the original image with the distribution conditioned on a distorted image, thereby down-weighting tokens that are more likely to arise from language priors or statistical biases rather than from the visual input. This baseline is relevant because many of our examples require the model to remain grounded in the reliable image input. However, we note that VCD is primarily designed to improve visual grounding without identifying which modality or input component is trustworthy under conflicts. It therefore provides a useful comparison for cases where visual grounding is beneficial, but it is not a general method to handle conflicts, especially when text should be trusted.

    Our implementation uses the hyperparameters as 
    $$
    \alpha=1,\quad \beta=0.1,\quad \gamma=0.1,\quad T=500.
    $$
    The diffusion process uses $\gamma$ for its noise schedule and the distorted-image signal coefficient was $(1-\gamma)^{T/2}$. Decoding used
    
    $$
    \ell_{\mathrm{VCD}}
    =(1+\alpha)\ell_{\mathrm{original}}
    -\alpha\ell_{\mathrm{distorted}},
    $$
    together with the adaptive plausibility constraint, which prevents contrastive decoding from promoting tokens that are implausible under the original image. Only tokens satisfy
    $$
    p(y_t\mid v,x)\geq
    \beta\max_w p(w\mid v,x)
    $$
    are retained.

    \paragraph{Modality Preference Steering (MPS)}
    
    MPS~\cite{zhang2025evaluating} is introduced in the same work as the MC$^2$ benchmark. MPS is particularly aligned with our setting, as it is designed to control an MLLM's preference between visual and textual evidence when the two sources provide competing signals. It first probes the model with paired instructions that encourage reliance on either the visual context or the textual context, estimates a modality-preference direction in hidden-state space, and then adds a scaled steering vector during inference to shift the model toward the desired modality.

    In our implementation, a modality-preference vector was computed offline from 500 MC$^2$ samples. Each example was presented twice using image- and text-guided conflict instructions in this order: text context, modality instruction, question, and general direct-answer guidance. The modality instructions are as follows

\begin{icmlpromptbox}
\begin{verbatim}
In case there is an inconsistency 
between the text context and the 
image content, you should follow 
the image content rather than the 
text context.
\end{verbatim}
\end{icmlpromptbox}
\begin{icmlpromptbox}
\begin{verbatim}

In case there is an inconsistency 
between the text context and the 
image content, you should follow 
the text context rather than the 
image content.
\end{verbatim}
\end{icmlpromptbox}

    At the selected decoder layer, the final non-padding prompt-token state was recorded. The text direction was

    $$
    v=\frac{1}{N}\sum_i(h_i^{T}-h_i^{V}),
    $$
    and was normalized by the mean text-state norm divided by $||v||_2$. Steering to text adds the resulting vector to every token state, whereas steering to image subtracts it. The zero-based intervention layers were 22 for Qwen2.5, 28 for Qwen3, 22 for LLaVA-OV, and 16 for InternVL3. As the steering directly applies to the hidden states, the performance and response generation quality are sensitive to the choice of intervention scales. 
    We use the scale of 0.25 for MMCD, MC$^2$, CLASH, and CorrelationQA and 0.75 for PhD subsets for Qwen3-VL-8B to ensure stable answer quality. The remaining models, including Qwen2.5-VL-7B, LLaVA-OneVision-7B, and InternVL3-8B, follow the standard scale of 1.0 as in~\cite{zhang2025evaluating}.

\paragraph{Ours}
    Our main experiments in Tables 3 and 4 apply the arbitration head localization for every MLLM on 10 samples, each containing an image, a conflicting text, a question, and image-/text-supported answers. The top-30 heads are applied for the intervention on all the benchmarks with an intervention strength $\alpha$ of 30. The $\epsilon$ in Equation 5 is 1e-12. For the adaptive steering in Section 4.4, we use a fixed seed-42 sample of 200 MC$^2$, supplemented with the aligned text to form 200 aligned/conflicting pairs. The threshold $\tau$ obtained on the 200 samples is 0.801, with 83.0\% conflict accuracy, 74.0\% aligned accuracy, 78.5\% balanced accuracy, and AUC 0.8517. We use $\alpha_\text{max}$ of 50 and $\lambda$ of 0.05 for the intervention strength $\alpha$ choice in Equation 7.

\subsection{MC$^2$ Aligned Text Generation}
For the adaptive steering experiment in Section 4.4, we supplemented the text that yields the image-supported answers, but in a similar format to the conflicting text in the original MC$^2$. Specifically, we use gpt-4.1-mini and input the original conflicting context, text-supported answer, and image-supported answer. We then rewrote statements associated with the text answer to be consistent with the vision answer while preserving unrelated wording, detail, style, and structure as closely as possible. The prompt for rewriting is as follows.

We show an example of the aligned text below, compared to Example 2.
The experiments in Section 4.4 use a mixture of the aligned and conflicted input data, and the ratio of conflicted samples is controlled by a hyperparameter of 0, 0.2, 0.4, 0.6, 0.8, and 1.
\begin{icmlpromptbox}
\begin{verbatim}
You rewrite a context so its relevant 
evidence agrees with a target answer.
Replace every statement, description, 
implication, synonym, or paraphrase 
semantically associated with the 
original text answer with content 
consistent with the vision answer.
Preserve all unrelated content as 
closely as possible, including 
wording, detail, tone, style, 
paragraphing, and structure. Resolve 
any nearby wording that would 
otherwise contradict the vision 
answer, but do not invent unrelated 
details. Return only the rewritten 
context in the requested structured 
format.
\end{verbatim}
\end{icmlpromptbox}

\subsection{CorrelationQA Image-supported Answer Generation}~\label{app:corrqa}
    The original CorrelationQA only provides the text-supported answers, and we supplement the image-supported answers by instructing gpt-4.1-mini to generate answers with only the image and the decomposed final question.
    The model was instructed to provide the most factually correct image-grounded answer in no more than two words, without explanation or negation. Each example allowed at most three API calls, with constraint-focused retry instructions when necessary. If all attempts violated the format, the final answer was retained in case the answer needed more than two words. The prompt to generate the image-supported answers is as follows:

\vqaexample
    {Example 7: MC$^2$ (aligned)}
    {figures/files/mc2.png}
    {On a sunny afternoon, a group of young men are gathered at the edge of a soccer field, engaged in an intense discussion about their upcoming match strategies. Among them, a dark-haired individual is running with a clipboard, jotting notes as he moves and occasionally gesturing to emphasize points in the conversation. Rather than standing still, he actively moves around, facilitating the tactical planning and ensuring everyone understands their roles for the game ahead.}
    {What is dark haired person doing?}
    {}
    {Running}

\begin{icmlpromptbox}
\begin{verbatim}
Answer the following question. An 
image is provided as additional 
context. Give the most factually 
correct answer from the image to 
the question. Return only the answer 
in no more than 2 words, without 
explanation. Do not use negation in 
the answer.
\end{verbatim}
\end{icmlpromptbox}

\subsection{Evaluation}

    Accuracy was reported independently against vision and text references whenever both were available. PhD reported vision accuracy only, including for text-directed methods. The default evaluator was the hybrid parser with an exact matcher and an LLM as the judge. For each generated response, we evaluate it against image-supported answer or text-supported answer in two runs.
    We first normalize case, punctuation, whitespace, and articles for the generated response. We accept a reference when its complete normalized token sequence appears contiguously in the generated response. For datasets with both image-/text-supported answers, we short-circuit when exactly one reference matches. If both or neither match, call gpt-4o-mini separately for the vision and text references. For PhD, the single-reference dataset, short-circuit on a positive match, and otherwise call the judge.
    
    The judge used temperature 0, structured boolean output, 128 maximum output tokens, and prompt version 1.5. It judged semantic equivalence and accepted synonyms, paraphrases, harmless specificity differences, and consistent additional details. Accuracy and extraction-failure rate were recorded separately for each ground-truth type. The prompt for the LLM judge is

\begin{icmlpromptbox}
\begin{verbatim}
You are an impartial answer-
equivalence judge. Decide whether 
a candidate answer is semantically 
equivalent to the provided reference 
answer for the given question. Judge 
meaning, not exact wording. Accept 
synonyms, paraphrases, equivalent 
numbers, harmless grammatical or 
formatting differences, and reasonable 
differences in specificity or detail. 
Accept the subset or superset of a 
ground-truth answer as long as any 
omitted or added detail does not 
materially change or contradict the 
core answer. Mark the result incorrect 
only if it contradicts or is 
semantically inconsistent with the 
ground-truth answer. Return the 
requested structured result only.
\end{verbatim}
\end{icmlpromptbox}

\section{Computational Efficiency}
    Experiments were conducted on a machine with NVIDIA driver 580.173.02 and three RTX A5000 GPUs, each with 24\,GB and each run used a single GPU.
    The main software dependencies included Python 3.10.20, PyTorch 2.12.0+cu130, Transformers 5.7.0, CUDA 13.0, and cuDNN 9.2.

    We evaluate the computational efficiency of different inference methods using end-to-end latency (s/request), time to first token (TTFT; s), decode latency per output token (s/token), token throughput (tokens/s), peak GPU memory (GiB), and output token count (tokens). End-to-end latency measures the wall-clock time from the beginning of an inference request until the decoded response is returned, including input preprocessing and any method-specific computation. TTFT measures the time from the start of generation to the first generated token, while decode latency measures the subsequent generation time normalized by the number of decoded tokens. Token throughput is computed as the total number of generated tokens divided by the total generation time, including prefill. Peak GPU memory corresponds to the maximum CUDA memory allocated during inference, including both the resident model and temporary inference tensors. Output token count measures the number of tokens generated per request.

    We report efficiency results on MC$^2$ and compute all metrics over unique inference batches to avoid double-counting the image- and text-reference evaluation rows corresponding to the same generation. End-to-end latency (s/request) and TTFT (s) are averaged across requests, decode latency (s/token) is pooled across generated tokens, token throughput (tokens/s) is computed from the total output tokens and generation time, and peak GPU memory (GiB) is taken as the maximum across inference batches. Output tokens are reported as the average number of generated tokens. For our method, we measure the main intervention's computational efficiency for comparison with other methods, while we note that the adaptive method requires additional unimodal inference for conflict detection. For methods that support bidirectional modality steering, we report the image-steering setting for a consistent comparison. 
    
    \begin{table*}[t]
\centering
\resizebox{0.97\linewidth}{!}{
\begin{tabular}{ll*{6}{c}}
\toprule
\textbf{Model} & \textbf{Method} & \textbf{End-to-end latency} & \textbf{TTFT} & \textbf{Decode latency} & \textbf{Token throughput} & \textbf{Peak GPU memory} & \textbf{Output tokens} \\
\midrule
\multirow{6}{*}{Qwen3-VL-8B}
& Base
& 0.232 & 0.134 & 0.037 & 14.705 & 16.500 & 3.162 \\

& CoT
& 4.857 & 0.138 & 0.036 & 26.898 & 16.502 & 130.129 \\

& VCD
& 0.637 & 0.270 & 0.172 & 4.873 & 16.601 & 2.933 \\

& CAP
& 0.225 & 0.136 & 0.038 & 13.837 & 16.502 & 2.884 \\

& MPS
& 0.245 & 0.136 & 0.039 & 14.622 & 16.500 & 3.314 \\

& Ours
& 0.275 & 0.148 & 0.050 & 12.300 & 16.556 & 3.146 \\

\midrule

\multirow{6}{*}{Qwen2.5-VL-7B}
& Base
& 0.383 & 0.189 & 0.027 & 20.558 & 15.594 & 7.492\\

& CoT
& 1.398 & 0.184 & 0.026 & 33.718 & 15.596 & 46.440\\

& VCD
& 2.123 & 0.365 & 0.211 & 4.397 & 15.730 & 9.192 \\

& CAP
& 0.400 & 0.182 & 0.026 & 22.817 & 15.594 & 8.701 \\

& MPS
& 0.738 & 0.181 & 0.026 & 29.982 & 15.594 & 21.578 \\

& Ours
& 0.491 & 0.194 & 0.036 & 18.237 & 15.654 & 8.545 \\

\midrule

\multirow{6}{*}{LLaVA-OV-7B}
& Base
& 0.678 & 0.612 & 0.028 & 4.003 & 15.722 & 2.630 \\

& CoT
& 0.707 & 0.642 & 0.028 & 3.711 & 15.724 & 2.546 \\

& VCD
& 2.398 & 1.262 & 0.675 & 1.112 & 16.363 & 2.617 \\

& CAP
& 0.678 & 0.613 & 0.028 & 3.982 & 15.724 & 2.620 \\

& MPS
& 0.662 & 0.601 & 0.028 & 3.745 & 15.722 & 2.400 \\

& Ours
& 0.960 & 0.875 & 0.041 & 2.708 & 19.301 & 2.539 \\

\midrule

\multirow{6}{*}{InternVL3-8B}
& Base
& 0.730 & 0.631 & 0.028 & 4.516 & 15.520 & 3.113 \\

& CoT
& 3.366 & 0.596 & 0.027 & 30.969 & 15.523 & 103.214 \\

& VCD
& 2.585 & 1.253 & 0.667 & 1.150 & 16.125 & 2.890 \\

& CAP
& 0.726 & 0.632 & 0.028 & 4.400 & 15.523 & 3.029 \\

& MPS
& 0.721 & 0.603 & 0.028 & 5.734 & 15.520 & 3.917 \\

& Ours
& 0.906 & 0.780 & 0.037 & 3.830 & 18.461 & 3.320 \\

\bottomrule
\end{tabular}
}
\caption{Efficiency comparison of methods across MLLMs.
}
\label{appd:tab:computing}
\end{table*}

    The results show that our method only yields limited inference-time overhead while maintaining output lengths comparable to standard inference. On Qwen3-VL-8B and Qwen2.5-VL-7B, our method increases end-to-end latency moderately relative to Base and CAP, while remaining substantially faster than VCD or MPS on Qwen2.5-VL-7B. On LLaVA-OV-7B and InternVL3-8B, the latency increase over Base and CAP is larger, but our method remains considerably faster than VCD. Across all four models, the number of output tokens produced by our method is close to that of Base and CAP, indicating that its additional latency arises from the inference-time intervention rather than longer generations. In contrast, VCD yields substantially higher TTFT and decode latency across all models, while CoT can lead to high end-to-end latency when it produces substantially longer responses. Peak GPU memory for our method remains close to Base on the two Qwen models, with larger increases on LLaVA-OV-7B and InternVL3-8B. Overall, our method provides effective modality arbitration with moderate computational overhead.

\section{Additional Results}

\subsection{Robustness of Modality Conflict Detection}
To assess the robustness of our conflict diagnostic, we compute cosine similarity after mean-centering each logit vector to remove sensitivity to uniform additive shifts. The balanced accuracy decreases only slightly, from 78.5\% with the original cosine similarity to 77.5\% with mean-centered cosine similarity. The conflicting accuracy decreases from 83\% to 81\%, while the aligned accuracy remains unchanged at 74\%. These results suggest that the observed conflict signal is largely robust to uniform additive shifts in the logits.

\begin{figure}
    \centering
    \includegraphics[width=0.9\linewidth]{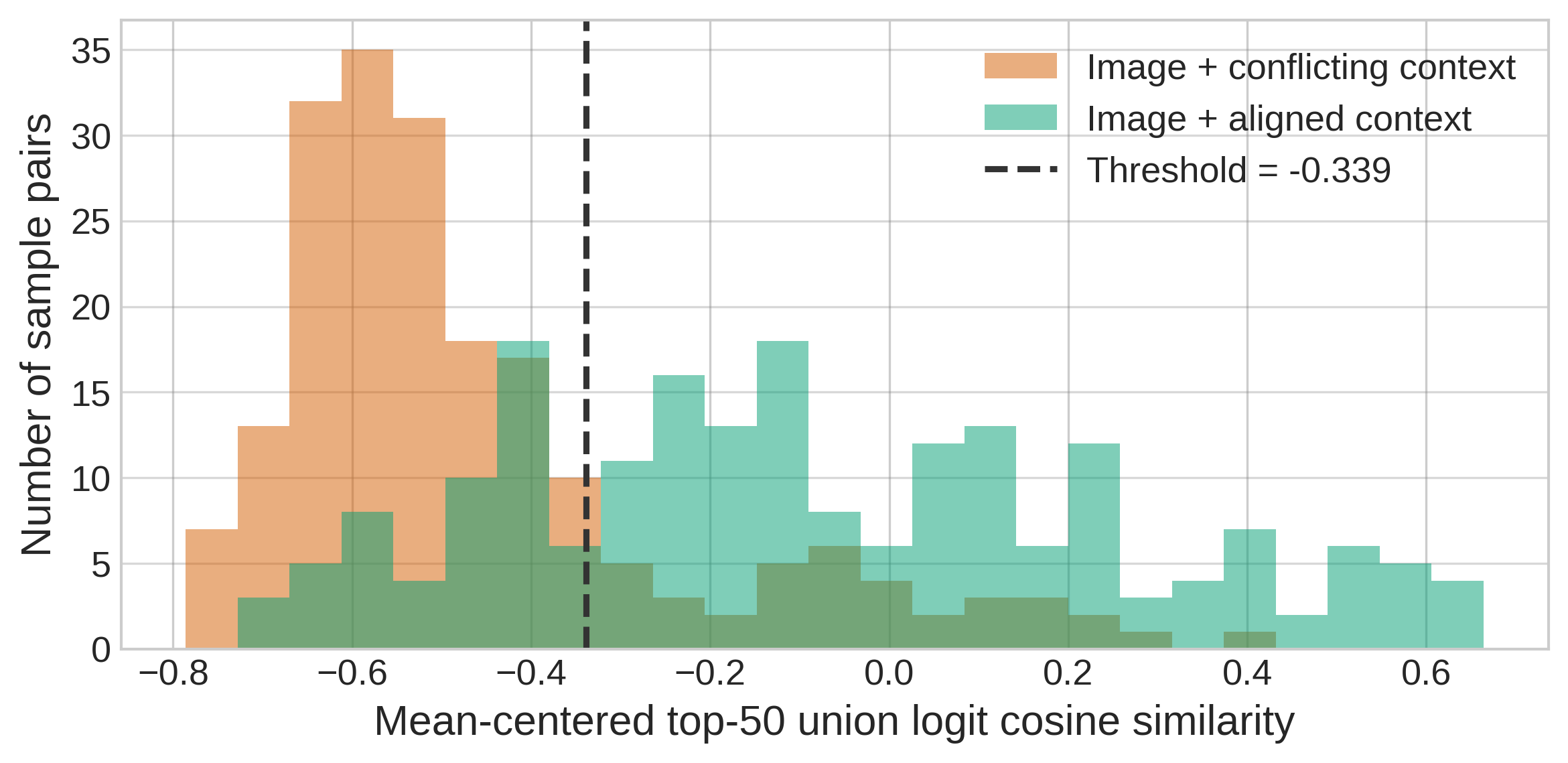}
    \caption{
    Robustness of modality conflict detection using centered logits. Top-50 union-logit cosine similarities after centering each logit vector, for samples paired with conflicting or aligned context. Lower similarity indicates stronger modality conflict. The dashed line marks the conflict-detection threshold calibrated to maximize accuracy on the calibration set.
    }
    \label{fig:threshold2}
    \vspace{-2mm}
\end{figure}

Further, for assessing sensitivity to calibration-set selection, we repeat threshold calibration using a different random seed of 43. The resulting threshold is 0.8057, yielding a balanced accuracy of 75.5\%, with 79\% conflicting accuracy and 72\% aligned accuracy. These results indicate that conflict detection remains effective under a different calibration-set selection, although performance varies modestly with the calibration seed.

\begin{figure}
    \centering
    \includegraphics[width=0.9\linewidth]{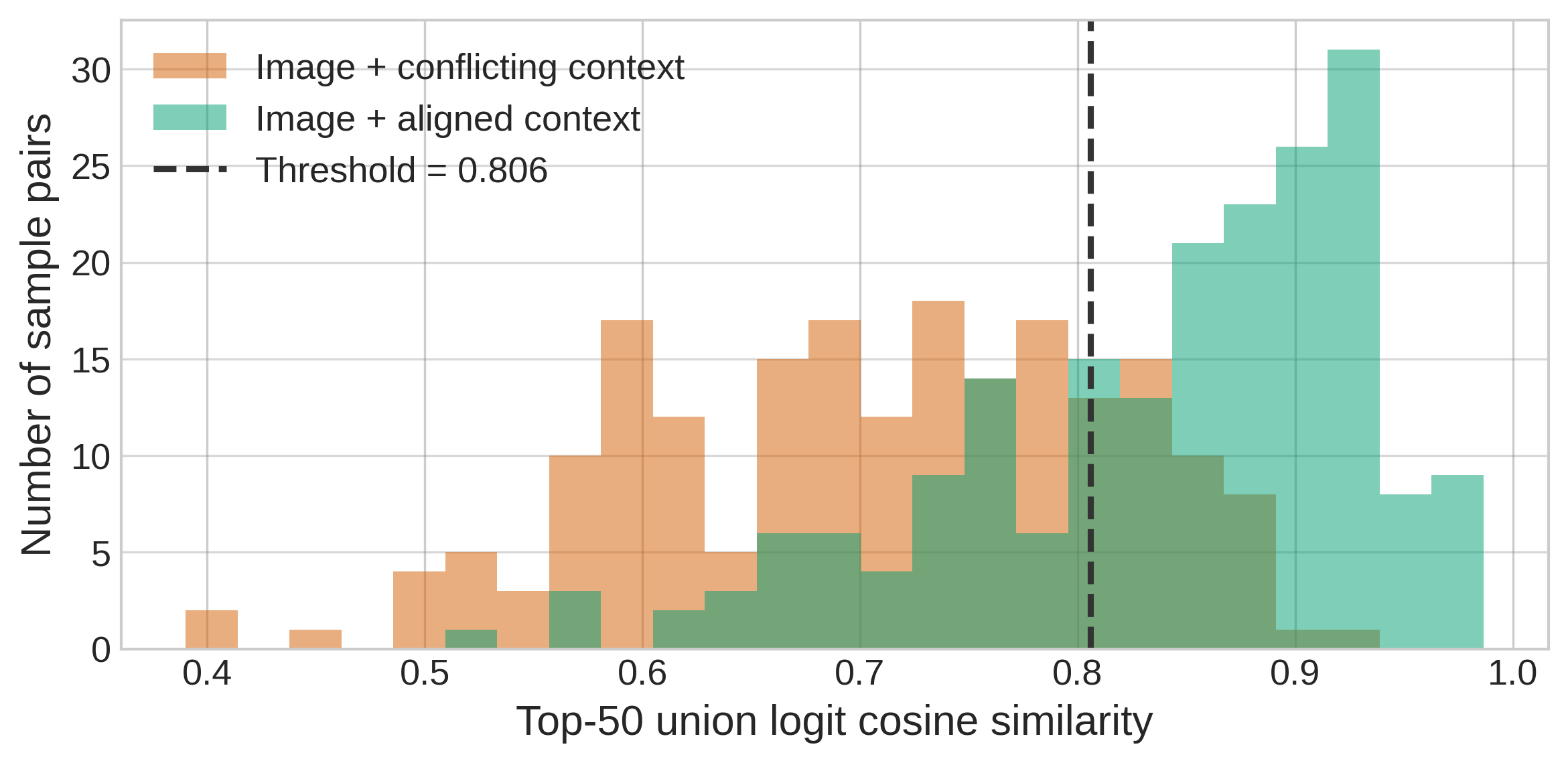}
    \caption{
    Robustness of modality conflict detection to calibration-set selection. Top-50 union-logit cosine similarities for samples paired with conflicting or aligned context, using a calibration set generated with random seed 43. The dashed line marks the detection threshold selected to maximize conflict-detection accuracy on the calibration set.
    }
    \label{fig:threshold3}
    \vspace{-2mm}
\end{figure}

\subsection{Visual Degradation Examples}
    Table 2 shows the performance of Qwen3-VL-8B when the image quality is degraded with Gaussian blurring. We include an example of MC$^2$ for Blur 1-3.

\begin{figure*}[t!]
    \centering
    \begin{subfigure}{0.24\linewidth}
        \includegraphics[width=\linewidth]{figures/files/mc2.png}
        \caption{Original image}
    \end{subfigure}
    \hfill
    \begin{subfigure}{0.24\linewidth}
        \includegraphics[width=\linewidth]{}
        \caption{Blur-1}
    \end{subfigure}
    \hfill
    \begin{subfigure}{0.24\linewidth}
        \includegraphics[width=\linewidth]{}
        \caption{Blur-2}
    \end{subfigure}
    \hfill
    \begin{subfigure}{0.24\textwidth}
        \includegraphics[width=\textwidth]{}
        \caption{Blur-3}
    \end{subfigure}
    \label{appd:fig:blur}
    \caption{Examples of Blur 1-3 on one MC$^2$ image.}
\end{figure*}

\subsection{Heatmaps for Other MLLMs}

    In addition to the heatmap visualization for Qwen3-VL-8B, we further examine the stability of arbitration-head localization on the other MLLMs as the calibration set size increases from \(n=3\) to \(n=10\). Across LLaVA-OneVision-7B, Qwen2.5-VL-7B, and InternVL3-8B, high-influence heads repeatedly emerge at similar layer–head locations across different sample sizes. Several prominent heads are already identifiable with only three samples and become increasingly stable as more calibration examples are used. Despite model-specific differences in the overall distribution of influence scores, the persistence of high-influence locations suggests that arbitration heads can be reliably localized from a small calibration set across different MLLM architectures.

\begin{figure*}
    \centering
    \includegraphics[width=0.8\linewidth]{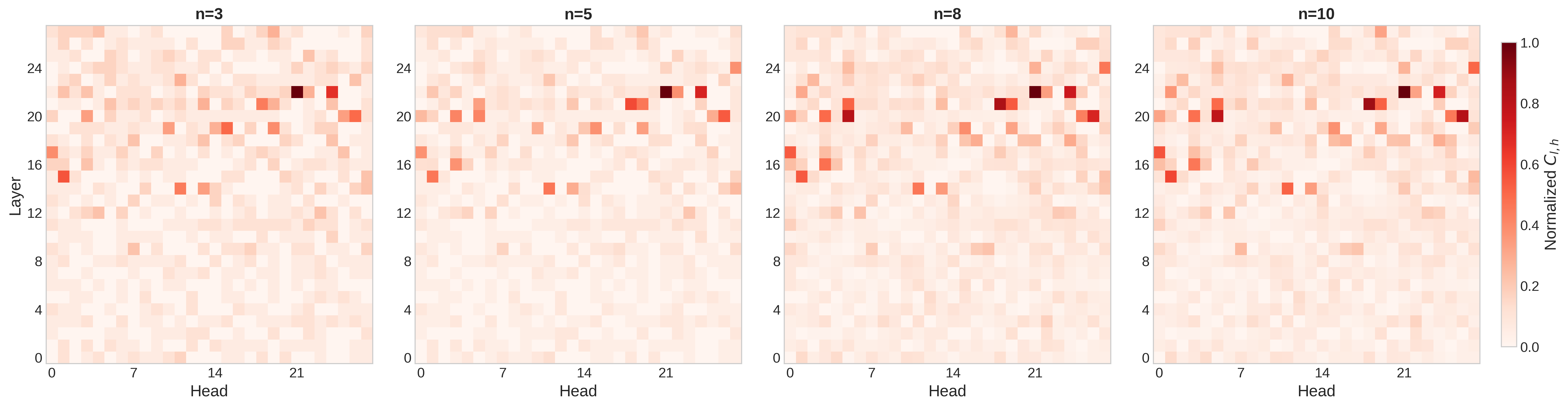}
    \caption{Stability of head-wise effects across different sample sizes for Qwen2.5-VL-7B.
    }
    \label{appd:fig:heatmap-qwen}
\end{figure*}

\begin{figure*}
    \centering
    \includegraphics[width=0.8\linewidth]{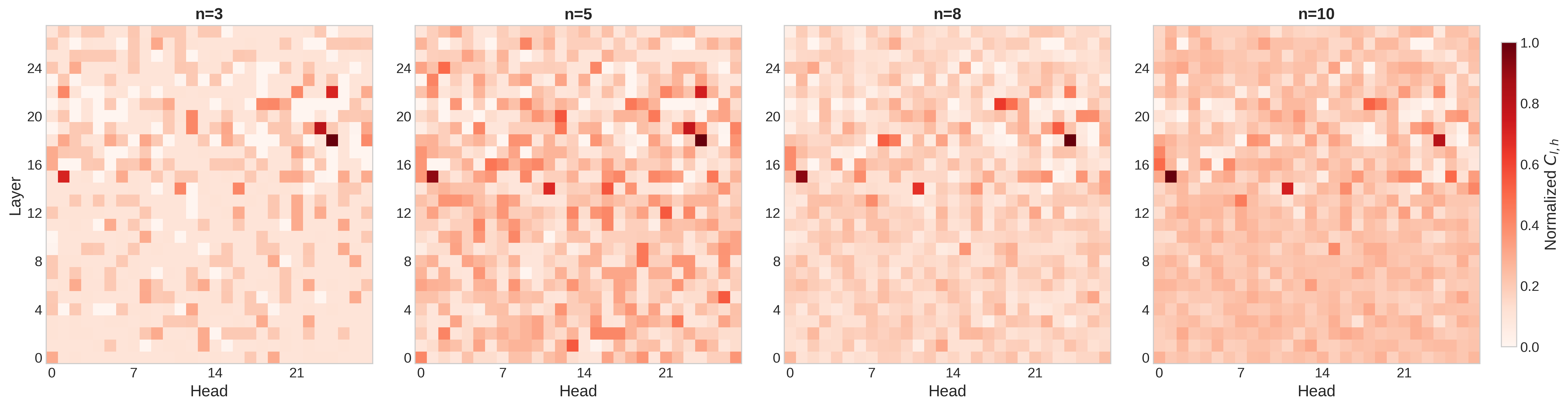}
    \caption{Stability of head-wise effects across different sample sizes for LLaVA-OneVision-7B.
    }
    \label{appd:fig:heatmap-llava}
\end{figure*}

\begin{figure*}
    \centering
    \includegraphics[width=0.8\linewidth]{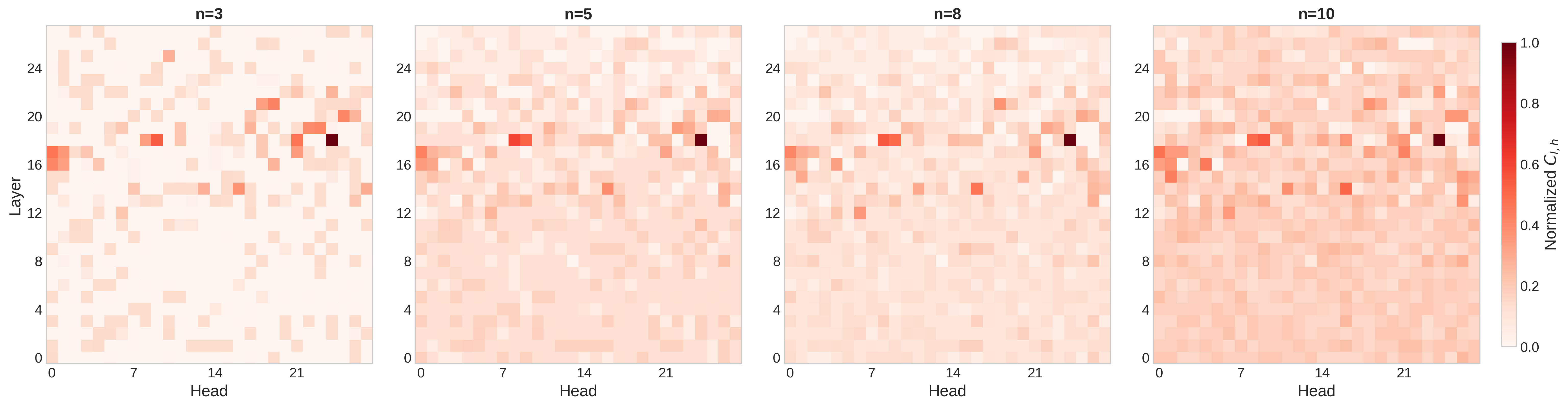}
    \caption{Stability of head-wise effects across different sample sizes for InternVL3-8B.
    }
    \label{appd:fig:heatmap-internvl}
\end{figure*}

\subsection{Robustness to Visual Degradation}
    We further examine modality arbitration performance as the reliability of visual evidence decreases by progressively blurring the input image. Base, CoT, and VCD increasingly shift toward the text-supported answer as image quality decreases, demonstrating that modality arbitration is sensitive to the relative strength of cross-modal evidence. When steering toward vision, our method consistently preserves higher vision-answer accuracy than the bidirectional baselines across degradation levels, although its effectiveness gradually decreases under severe blur. Conversely, when steering toward text, our method maintains consistently high text-answer accuracy across all degradation levels. These results demonstrate that the proposed intervention provides robust bidirectional control over modality reliance even as the relative reliability of the visual and textual evidence changes.

\begin{figure*}
    \centering
    \includegraphics[width=0.8\linewidth]{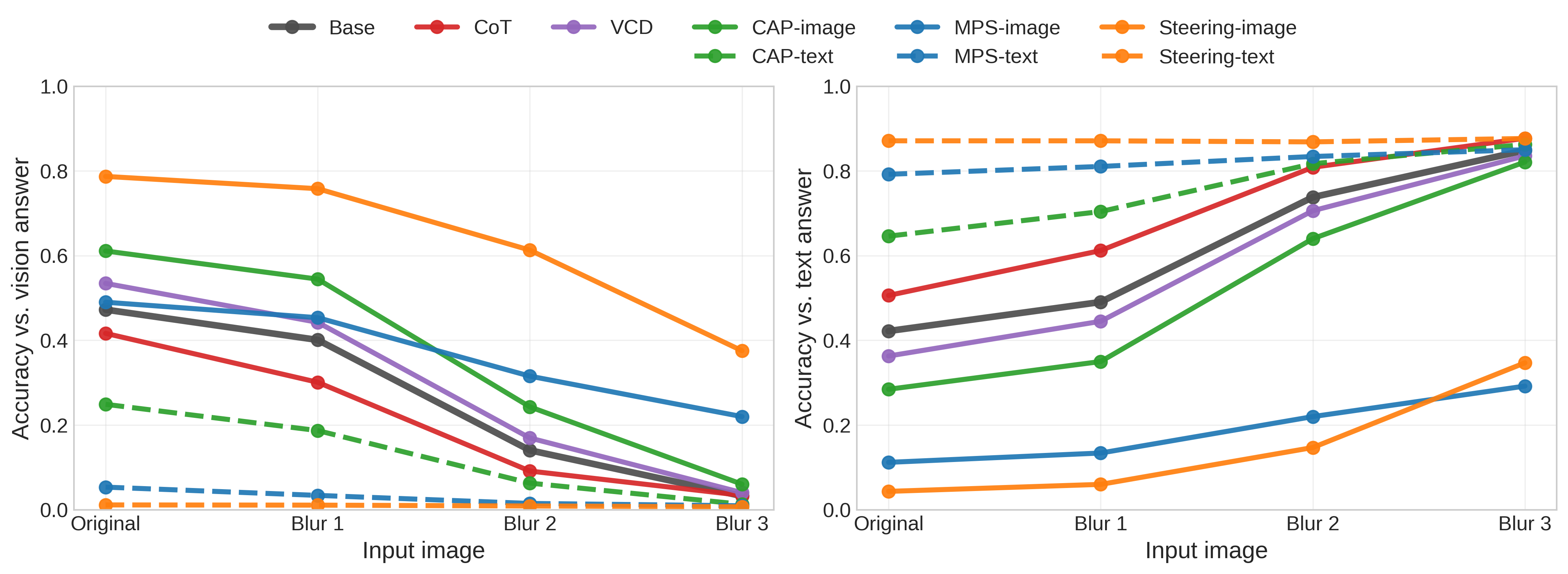}
    \caption{
    Robustness to visual degradation.
    }
    \label{fig:blurring}
    \vspace{-2mm}
\end{figure*}

\section{Detailed Discussion}\label{app:discussion}
Our work highlights several limitations and opportunities for future work. 
First, as we find that heads identified using only 10 samples generalize well across the evaluated benchmarks, it remains unclear whether the same patterns transfer to substantially different domains or real-world tasks. 
Second, our evaluation uses a simplified modality-conflict setup in which the modality to prioritize is externally specified. In realistic settings, both modalities may provide useful yet partially conflicting evidence, requiring the model to integrate complementary information rather than rely exclusively on one modality. Our current evaluation does not fully capture such cases, and benchmarking our intervention in settings where unimodal routing is insufficient remains an important direction for future work. Automatically assessing modality reliability and identifying which evidence to trust are also open challenges.
Following this, the aligned MC$^2$ evaluation in our work measures whether adaptive steering retains the benefit of multimodal inference when both modalities support the same answer, but it does not test examples that require combining distinct evidence from both modalities. Preservation of useful competing-modality information under partial conflict therefore remains to be evaluated.
Finally, we introduce an adaptive steering extension that detects conflicts by comparing image-only and text-only predictions. While this approach demonstrates practical applicability in settings where conflicts are sparse, it introduces additional inference cost for conflict detection. Developing more efficient conflict signals that can be extracted directly from a single multimodal forward pass would be a promising direction for future work.

\end{document}